%% file: camera_ready.tex
\documentclass{article}
    \PassOptionsToPackage{numbers, compress}{natbib}
\usepackage[final]{neurips_2022}

\usepackage[utf8]{inputenc} % allow utf-8 input
\usepackage[T1]{fontenc}    % use 8-bit T1 fonts
\usepackage[pagebackref=true,breaklinks=true,colorlinks,bookmarks=false]{hyperref}
\usepackage{url}            % simple URL typesetting
\usepackage{booktabs}       % professional-quality tables
\usepackage{amsfonts}       % blackboard math symbols
\usepackage{nicefrac}       % compact symbols for 1/2, etc.
\usepackage{microtype}      % microtypography
\usepackage{xcolor}         % colors
\usepackage{graphicx}
\usepackage{caption}
\usepackage{amsmath}
\usepackage[capitalise]{cleveref}
\usepackage{amssymb}
\usepackage{pifont}
\usepackage{wrapfig}
\usepackage{comment}
\usepackage{adjustbox}
\usepackage{array}
\newcolumntype{R}[2]{%
    >{\adjustbox{angle=#1,lap=\width-(#2)}\bgroup}%
    l%
    <{\egroup}%
}
\usepackage{multirow}
\usepackage{enumitem}
\usepackage{algorithm}
\usepackage{algorithmicx}
\usepackage{algpseudocode}

\newcommand*\cellspacelimit[1]{\setlength{\cellspacetoplimit}{#1}\setlength{\cellspacebottomlimit}{#1}}
\usepackage{cellspace}
\usepackage{arydshln}
\usepackage{mathtools}
\usepackage[thinc]{esdiff}
\usepackage{subcaption}
\makeatletter
\def\adl@drawiv#1#2#3{%
        \hskip.5\tabcolsep
        \xleaders#3{#2.5\@tempdimb #1{1}#2.5\@tempdimb}%
                #2\z@ plus1fil minus1fil\relax
        \hskip.5\tabcolsep}
\newcommand{\cdashlinelr}[1]{%
  \noalign{\vskip\aboverulesep
          \global\let\@dashdrawstore\adl@draw
          \global\let\adl@draw\adl@drawiv}
  \cdashline{#1}
  \noalign{\global\let\adl@draw\@dashdrawstore
          \vskip\belowrulesep}}
\makeatother

\newcommand{\cmark}{\ding{51}}
\newcommand{\xmark}{\ding{55}}

\newcommand*\rot{\multicolumn{1}{R{90}{1em}}}% no optional argument here, please!

\title{UniCLIP: Unified Framework for \\ Contrastive Language--Image Pre-training}

\author{%
  Janghyeon Lee\thanks{Equal contribution. Alphabetical order.} \\
  LG AI Research \\
  \texttt{janghyeon.lee@lgresearch.ai} \\
  \And
  Jongsuk Kim$^{*}$\thanks{Work done during an internship at LG AI Research.} \\
  KAIST\\
  \texttt{jskpop@kaist.ac.kr} \\
  \And
  Hyounguk Shon$^{\dagger}$ \\
  KAIST \\
  \texttt{hyounguk.shon@kaist.ac.kr} \\
  \And
  Bumsoo Kim \\
  LG AI Research \\
  \texttt{bumsoo.kim@lgresearch.ai} \\
  \And
  Seung Hwan Kim \\
  LG AI Research \\
  \texttt{sh.kim@lgresearch.ai} \\
  \And
  Honglak Lee \\
  LG AI Research \\
  \texttt{honglak@lgresearch.ai} \\
  \And
  Junmo Kim \\
  KAIST \\
  \texttt{junmo.kim@kaist.ac.kr} \\
}

\begin{document}

\maketitle

\input{abstract}

\input{introduction}

\section{Methods}
\label{method}
\begin{figure}[t]
    \centering
    \includegraphics[width=\textwidth]{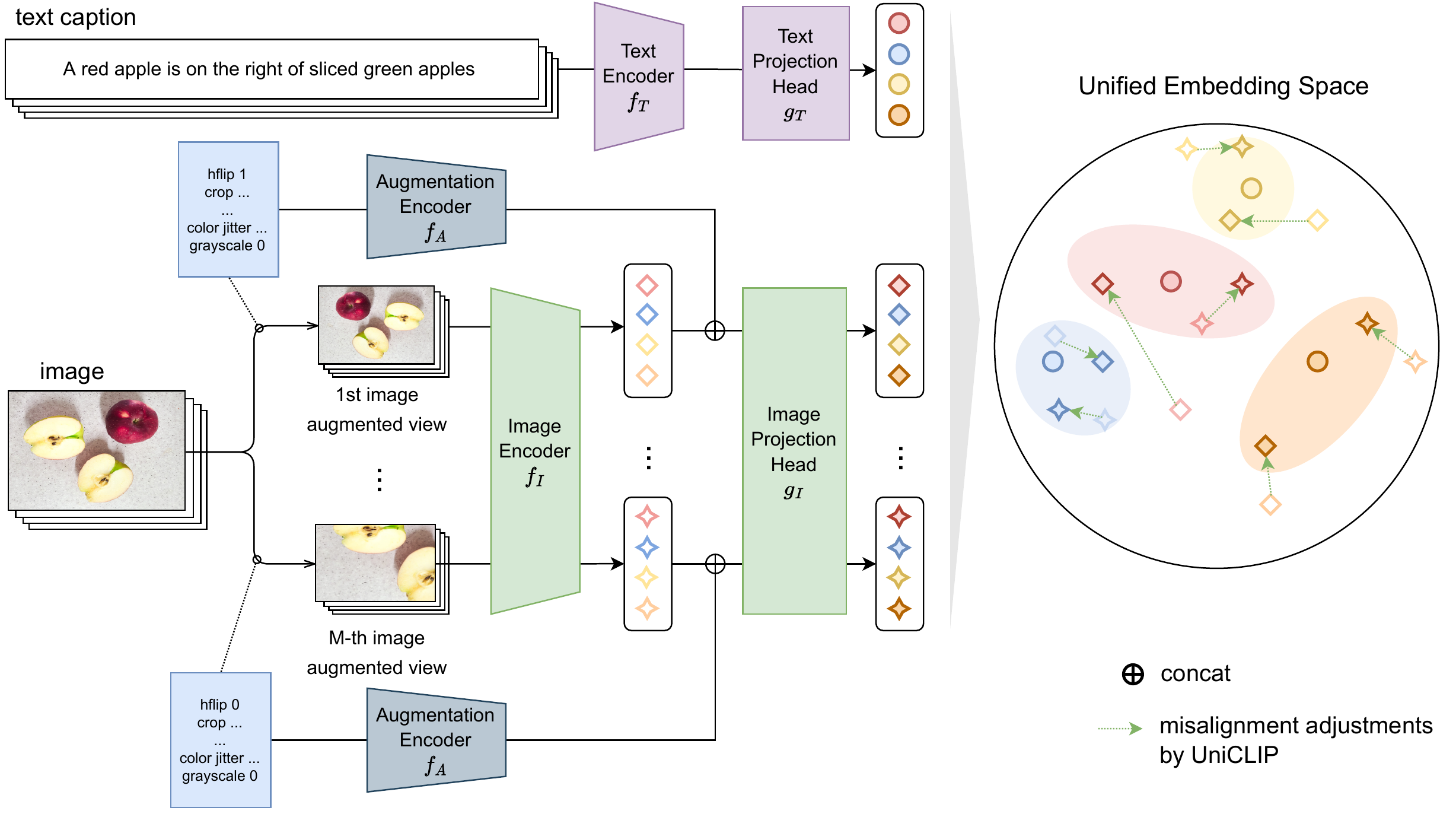}
    \caption{Overview of the UniCLIP framework.}
    \label{fig:overview}
\end{figure}

The UniCLIP architecture (Figure~\ref{fig:overview}) consists of an augmentation encoder $f_A$, an image encoder $f_I$, a text encoder $f_T$, and corresponding projection heads $g_I$ and $g_T$.
$f_I$ encodes an image to an augmentation-agnostic image representation and then $g_I$ outputs an augmentation-aware image embedding.
For text caption data, $f_T$ and $g_T$ produce text embeddings on the \emph{same} embedding space as the image embedding space.
Image and text representations are learned by our multi-positive NCE loss with domain-dependent similarity scores measured on the unified embedding space.
Each element of our method is described in detail in the following sections.

\subsection{Architecture}
\label{method:arch}

\paragraph{Augmentation Encoder}
To enable an augmentation instruction $\mathcal{A}$ to be used as an input to a network, we first describe it as a real vector containing information about how much each basic transformation in $\mathcal{A}$ is applied to data. For example, image augmentations that frequently appear in contrastive learning can be converted to real vectors as follows:
\begin{itemize}
    %\small
    \item \textit{Crop \& Resize}: 
    A \texttt{RandomResizedCrop} augmentation is encoded to a four-dimensional vector of $(x, y, w, h)$, where $(x,y)$ is the top left corner coordinate of a cropped image and $(w, h)$ is the size of the cropped image, in a normalized coordinate system
    (\textit{i.e.}, the top left corner of the original image is $(0,0)$ and the bottom right corner is $(1,1)$).
    
    \item \textit{Color Jitter}: 
    As a \texttt{ColorJitter} augmentation changes the brightness, contrast, saturation, and hue of an image, this augmentation is encoded to a four-dimensional vector consisting of the changes in those four factors.
    
    \item \textit{Gaussian Blur}:
    A \texttt{GaussianBlur} augmentation is encoded to the standard deviation of its Gaussian blurring kernel.
    
    \item \textit{Horizontal Flip}: 
    A \texttt{RandomHorizontalFlip} augmentation is encoded to $1$ if an image is actually flipped and $0$ otherwise.
    
    \item \textit{Grayscale Convert}: 
    A \texttt{RandomGrayscale} augmentation is encoded to $1$ if an image is actually converted to grayscale and $0$ otherwise.
\end{itemize}

If an image augmentation $\mathcal{A}$ is composed of all five augmentations described above, $\mathcal{A}$ will be first encoded to an $11$-dimensional vector according to the above rules and then pass through an MLP to obtain the augmentation embedding $f_A(\mathcal{A})$.
Note that $f_A(\mathcal{A})$ will be different for each forward and each sample because of the randomness of the augmentation.

\paragraph{Image Encoder \& Image Projection Head}
For the model to learn how to adjust for image--text misalignment caused by image augmentations, the image encoder or projection head must take the augmentation information as input.
However, the encoder cannot fully benefit from augmented data if it knows which augmentation was applied to the image.
For example, when the encoder is trained with horizontal flip augmentation, and if it takes an augmented image and an flag of whether the image is flipped or not as input of the form (image, not flipped flag) or (flipped image, flipped flag), then the encoder may exhibit undesirable behavior when it has to encode (flipped image, not flipped flag) from some downstream task, since the encoder was not trained on this kind of data, which means that the model has lost some generalization ability.
Therefore, the image encoder must be augmentation-\emph{agnostic} and the image projection head must be augmentation-\emph{aware}.
In this way, the encoder can fully enjoy the benefits of data augmentation and generalizes better, while the projection head is still able to correct inter-domain misalignments caused by the augmentations.

To make image representations augmentation-agnostic and image embeddings augmentation-aware, the augmentation information is provided only to the projection head, whereas the encoder only sees the augmented image without knowing which augmentation has been applied.
Therefore, for an image $x$, the image encoder $f_I$ takes an augmented image $\mathcal{A}(x)$ as input to get an augmentation-agnostic image representation $h = f_I(\mathcal{A}(x))$.
Then, an augmentation-aware image embedding $z = g_I(f_I(\mathcal{A}(x)), f_A(\mathcal{A}))$ in the unified embedding space is obtained from the image representation $h$ and the augmentation embedding $f_A(\mathcal{A})$ by the image projection head $g_I$.
We adopt ViT (Vision Transformer)~\cite{dosovitskiy2021an} as the image encoder $f_I$ with learnable positional embeddings and the image projection head $g_I$ is composed of three residual blocks.
The last activation value of the \texttt{[cls]} token is used as the image representation $h$.

\paragraph{Text Encoder \& Text Projection Head}
A raw text is first tokenized by byte pair encoding and wrapped with a start token and an end token, resulting in a tokenized text $x$.
Any text augmentation method can also be applied here as in the case of the image embedding, but we do not create multiple augmented views for a text as we found it not very helpful.
So, the text representation $h=f_T(x)$ and the text embedding $z=g_T(f_T(x))$ in the unified latent space are obtained without any augmentation embedding.
We use Transformer~\cite{vaswani2017attention} for the text encoder $f_T$ with learnable positional embeddings and a linear layer for the text projection head $g_T$.
The last activation value of the start token is used as the text representation $h$.

\subsection{Contrastive Loss Functions for Multiple Positive Pairs}
\label{method:loss}
Contrastive loss functions can be classified according to the number of positive and negative pairs taken by the loss for one data point. For example, triplet loss~\cite{schroff2015facenet} takes only a single positive pair and a single negative pair, $N$-pair loss~\cite{sohn2016improved} and InfoNCE loss~\cite{van2018representation} take a single positive pair and multiple negative pairs, and MIL-NCE loss~\cite{miech2020end} and SupCon loss~\cite{khosla2020supervised} take multiple positive pairs and multiple negative pairs. As there are multiple positive pairs in our unified framework, we first review MIL-NCE loss and SupCon loss functions and discuss their drawbacks.

For an $i$-th embedding $z_i$ in a batch of embeddings $\{z_i\}_i$, let $P_i$ be the set of all positive sample indices of the $i$-th sample excluding $i$ itself and $N_i$ be the set of all negative sample indices of the $i$-th sample.
\begin{align}
P_i &= \{j | (z_i, z_j) \text{~is a positive pair and~} j \neq i\} \label{eq:pos} \\
N_i &= \{j | (z_i, z_j) \text{~is a negative pair}\}
\end{align}
A similarity score between the $i$-th and $j$-th embedding is denoted by $s_{i,j} > 0$.
A contrastive loss function will try to maximize the similarity scores of positive pairs, while minimize the similarity scores of negative pairs.
For example, if there is only one positive sample for each sample in a batch, say $P_i = \{ p_i \}$, then InfoNCE loss~\cite{van2018representation} or NT-Xent loss~\cite{chen2020simple} for the $i$-th sample can be described by
\begin{equation}\label{eq:infonce}
\mathcal{L}_i^\text{InfoNCE} = -\log \frac{s_{i,p_i}}{s_{i,p_i} + \sum_{n \in N_i} s_{i,n}} .
\end{equation}

\paragraph{MIL-NCE Loss}
MIL-NCE loss~\cite{miech2020end} for the $i$-th embedding is defined by
\begin{equation}~\label{eq:mil}
\mathcal{L}_i^\text{MIL-NCE} = -\log \frac{\sum_{p \in P_i} s_{i,p}}{\sum_{p \in P_i} s_{i,p} + \sum_{n \in N_i} s_{i,n}} .
\end{equation}
The MIL-NCE loss function is configured to maximize the sum of all positive pair similarity scores $\sum_{p \in P_i } s_{i, p}$ and minimize the sum of all negative pair similarity scores $\sum_{n \in N_i} s_{i,n}$.
However, hard positive pairs cannot receive enough gradients from $\mathcal{L}_i^\text{MIL-NCE}$ when there are easy positive pairs whose similarity scores are sufficiently large to dominate the numerator and denominator, as the MIL-NCE loss compares negative pairs with the sum of positive scores $\sum_{p \in P_i } s_{i, p}$ only, not each positive pair $s_{i,p}$ individually.
For some $q \in P_i$, the gradient from $\mathcal{L}_i^\text{MIL-NCE}$ to $s_{i,q}$ is
\begin{equation}
\frac{\partial\mathcal{L}_{ i}^\text{MIL-NCE}}{\partial s_{i,q}} = -\frac{\sum_{n \in N_i } s_{i, n}}{\left(\sum_{p \in P_i } s_{i,p}\right) \left(\sum_{p \in P_i } s_{i,p}+\sum_{n \in N_i} s_{i,n} \right)} ,
\end{equation}
therefore the gradient will vanish to zero when $\sum_{p \in P_i } s_{i, p}$ is already large because of easy positive pairs even if the positive pair's score $s_{i,q}$ is small.
In other words, \emph{easy positive pairs hinder the training of hard positive pairs} in MIL-NCE loss.
This problem will be more pronounced in our unified framework because hard positives and easy positives frequently coexist with supervisions from intra-domain and inter-domain.

\paragraph{SupCon Loss}
SupCon loss~\cite{khosla2020supervised} for the $i$-th embedding is described by
\begin{equation}\label{eq:supcon}
\mathcal{L}_{i}^\text{SupCon} = \mathbb{E}_{p \in P_i} \left[ - \log \frac{s_{i, p}}{\sum_{p' \in P_i } s_{i,p'}+\sum_{n \in N_i} s_{i,n}} \right] .
\end{equation}
In this case, each positive score $s_{i,p}$ is compared with the negative pairs, but the sum of the positive scores in the denominator still causes an undesirable side effect.
For an easy positive pair with a large similarity score, it can be possible to decrease the loss by decreasing its score and so the denominator.
For $q \in P_i$, 
\begin{equation}
\frac{\partial\mathcal{L}_{ i}^\text{SupCon}}{\partial s_{i,q}} = \frac{s_{i,q} - \frac{1}{|P_i|} \left( \sum_{p \in P_i} s_{i,p} + \sum_{n \in N_i} s_{i,n} \right)
}{s_{i,q} \left( \sum_{p \in P_i} s_{i,p} + \sum_{n \in N_i} s_{i,n} \right)} ,
\end{equation}
so hard positives would be trained better than MIL-NCE loss because of a relatively large update by the $s_{i,q}$ term in the denominator.
However, if we assume the sum of positive scores is much greater than the sum of negative scores, then
\begin{equation}
\frac{\partial\mathcal{L}_{ i}^\text{SupCon}}{\partial s_{i,q}} \propto s_{i,q} - \frac{1}{|P_i|} \left( \sum_{p \in P_i} s_{i,p} + \sum_{n \in N_i} s_{i,n} \right) \approx s_{i,q} - \mathbb{E}_{p \in P_i} \left[ s_{i,p} \right] .
\end{equation}
As gradient is not always negative, $\mathcal{L}_{i}^\text{SupCon}$ will try to decrease the similarity score of an easy positive pair $(z_i,z_q)$ since $s_{i,q}$ will be larger than the average positive score, instead of increasing or at least maintaining it. In other words, \emph{hard positive pairs hinder the convergence of easy positive scores} in SupCon loss.

\paragraph{Multi-positive NCE Loss}
As the sum of the positive scores in the denominator causes easy and hard positive pairs to interfere with each other, we can just use a multi-positive version of InfoNCE loss to make each positive pair independently contribute to the loss as follows.
\begin{equation}\label{eq:vanila}
\mathcal{L}_{i}= \mathbb{E}_{p \in P_i} \left[- \log \frac{s_{i, p}}{s_{i, p}+\sum_{n \in N_i} s_{i, n}} \right]
\end{equation}
As can be seen in the gradient
\begin{equation}
\frac{\partial \mathcal{L}_{i}}{\partial s_{i,q}} = -\frac{\sum_{n \in N_i} s_{i, n}}{|P_i| s_{i, q}  \left(s_{i, q}+\sum_{n \in N_i} s_{i, n}\right)} ,
\end{equation}
hard positive samples can be trained with sufficiently large update from the $s_{i,q}$ term in the denominator, and the decreasing easy positive pair similarity problem does not occur as the gradient is always negative.

With this multi-positive version of InfoNCE loss, we reconsider excluding $i$ from the positive set $P_i$ in Equation~\ref{eq:pos}.
If a contrastive loss can handle multiple positive pairs, then there is no reason to exclude the trivial pair $(z_i, z_i)$ from the loss definition.
Since $z_i$ is most similar to $z_i$ itself, the trivial pair must be also utilized as a strong positive pair, which will result in 
\begin{equation} \label{eq:ours_noweight}
\mathcal{L}_{i} = \mathbb{E}_{p \in P_i \cup \{i\}} \left[ - \log \frac{s_{i, p}}{s_{i, p}+\sum_{n \in N_i} s_{i, n}} \right] .
\end{equation}

Here, we propose a multi-positive NCE loss for our unified contrastive learning framework called MP-NCE loss, which is a weighted version of Equation~\ref{eq:ours_noweight} defined as
\begin{equation}\label{eq:ours}
\mathcal{L}_{i}^\text{MP-NCE}= \mathbb{E}_{p \in P_i \cup \{i\}} \left[ - w_{\mathcal{D}(i, p)}  \log \frac{s_{i, p}}{s_{i, p}+\sum_{n \in N_i} s_{i, n}} \right] ,
\end{equation}
where $\mathcal{D}(i, p)$ indicates the domain combination from which the $i$-th and $p$-th data were sampled,
and $w_{\mathcal{D}(i, p)}$ is a domain-specific balancing hyperparameter which makes each inter-domain and intra-domain supervision equally contributes to the loss.
For example, when we use three augmented views of an image and one corresponding text for each original image--text pair from dataset, there are a total of $9N$ image--image positive pairs, $6N$ image--text positive pairs, and $N$ text--text positive pairs in a batch, so $w_{\mathcal{D}(i, p)}$ is set to $1/9$, $1/6$, $1$ if $(z_i,z_p)$ is an image--image pair, image--text pair, text--text pair, respectively.

Although we have proposed MP-NCE loss in a multi-positive setting, one should consider using MP-NCE loss even in single positive settings, such as image self-supervised contrastive learning, by treating a trivial pair $(z_i, z_i)$ as positive as well since MP-NCE involves negligible computational overhead compared to backbone networks.

\subsection{Domain-Dependent Similarity Score}

In SimCLR~\cite{chen2020simple} and CLIP~\cite{radford2021learning}, the similarity score $s_{i,j}$ between the $i$-th embedding $z_i$ and $j$-th embedding $z_j$ is defined by
\begin{equation}
s_{i,j} = \exp \left( \frac{1}{\tau} \cdot \frac{z_i^\top z_j}{\|z_i\| \|z_j\|} \right) ,
\end{equation}
where $\tau$ is a positive real number, usually smaller than $1$.
As the cosine similarity of two embeddings cannot have a value outside the interval $[-1, 1]$, the cosine similarity is divided by the temperature $\tau$ to extend its range.
$\tau$ can be a pre-defined hyperparameter, or can rather be a learnable parameter allowing the model to choose an appropriate scale for the convergence of a contrastive loss.

To classify an input pair $(z_i, z_j)$ as positive or negative, we can define a threshold $b$ and classify it as positive if the cosine similarity between $z_i$ and $z_j$ is greater than $b$, and negative otherwise.
We may absorb this threshold $b$ into the similarity score as an offset like
\begin{equation} \label{eq:sim_th}
s_{i,j} = \exp \left( \frac{1}{\tau} \left(\frac{z_i^\top z_j}{\|z_i\| \|z_j\|} - b \right) \right) ,
\end{equation}
and expect that the optimal threshold will be learned by the model, as in the case of the temperature.
Note that the temperature will amplify the score if the cosine similarity is greater than $b$ otherwise reduce it, so Equation~\ref{eq:sim_th} is a reasonable similarity measure with which the threshold can be treated as a decision boundary for the binary classification problem.
However, unfortunately, the offset $b$ does not contribute to InfoNCE loss (Equation~\ref{eq:infonce}) at all since $b$'s in the numerator and denominator cancel out as
\begin{equation} \label{eq:add_offset}
\mathcal{L}_i^\text{InfoNCE} = -\log \frac{s_{i,p_i}}{s_{i,p_i} + \sum_{n \in N_i} s_{i,n}} = -\log \frac{\exp (b/\tau) s_{i,p_i}}{\exp (b/\tau) s_{i,p_i} + \sum_{n \in N_i} \exp (b/\tau) s_{i,n}}
\end{equation}
for any $\tau$ and $b$, which means ${\partial \mathcal{L}_i^\text{InfoNCE}}/{\partial b}$ is always zero.

On the other hand, when data pairs are sampled from multiple domains as in our unified framework, the threshold can be different depending on whether the sampled data pair is an intra-domain pair or an inter-domain pair, as it would be easier to classify intra-domain positive pairs than inter-domain positive pairs in general.
This motivates us to introduce domain-specific temperature $\tau_{\mathcal{D}(i,j)}$ and offset $b_{\mathcal{D}(i,j)}$, and propose a domain-dependent similarity score
\begin{equation}\label{eq:newsim}
s_{i,j} = \exp \left( \frac{1}{\tau_{\mathcal{D}(i,j)}} \left(\frac{z_i^\top z_j}{\|z_i\| \|z_j\|} - b_{\mathcal{D}(i,j)} \right) \right) .
\end{equation}
For image--text unified contrastive learning, we have three possible domain combinations, so there will be three different temperatures and three offsets respectively for image--image pairs, image--text pairs, and text--text pairs.

With the proposed domain-dependent similarity score (Equation~\ref{eq:newsim}) and MP-NCE loss (Equation~\ref{eq:ours}), the offsets are no longer cancelled out as negative pairs are sampled from multiple different domains.
Specifically, because any real number can be added to the cosine similarity term as in Equation~\ref{eq:add_offset} without changing the loss function, the offsets lose only $1$ intrinsic dimension and thus the model is able to learn \emph{relative} thresholds.
In other words, it is now possible to learn the domain-specific offsets so that we can expect the offset of an easier domain combination to be greater than that of harder one.

\section{Experiments}
\label{exp}
\begin{table}[t]
  \caption{ Zero-shot image classification performance and linear probing performance on 11 downstream datasets. $^\dagger$Results reported in the original paper.}
  \label{exp:main}
    \centering
    \begin{adjustbox}{max width=\textwidth}
    \begin{tabular}{lcccccccccccccc}
        \toprule
        Method & \shortstack{Pre-train \\ dataset} & \rot{Pets} & \rot{CIFAR-10} & \rot{CIFAR-100} & \rot{SUN397} & \rot{Food-101} & \rot{Flowers} & \rot{Cars} & \rot{Caltech-101} & \rot{Aircraft} & \rot{DTD} & \rot{ImageNet} & \rot{\textbf{Average}} \\
        \midrule
        \small\textit{Zero-shot classification:}\\
        CLIP-ViT-B/32 & YFCC15M & 19.4 & 62.3 & 33.6 & 40.2 & 33.7 & 6.3 & 2.1 & 55.4 & 1.4 & 16.9 & 31.3 & 27.5\\
        SLIP-ViT-B/32 & YFCC15M & 28.3 & 72.2 & 45.3 & 45.1 & 44.7 & 6.8 & 2.9 & 65.9 & 1.9 & 21.8 & 38.3 & 33.9\\
        DeCLIP-ViT-B/32 & YFCC15M & 30.2 & 72.1 & 39.7 & \textbf{51.6} & 46.9 & 7.1 & \textbf{3.9} & 70.1 & 2.5 & \textbf{24.2} & 41.2 & 35.4\\
        UniCLIP-ViT-B/32 & YFCC15M & \textbf{32.5} & \textbf{78.6} & \textbf{47.2} & 50.4 & \textbf{48.7} & \textbf{8.1} & 3.4 & \textbf{73.0} & \textbf{2.8} & 23.3 & \textbf{42.8} & \textbf{37.3}\\
        \cdashlinelr{1-14}
        DeCLIP-ResNet50$^\dagger$~\cite{li2022supervision} & Open30M & - & - & - & - & - & - & - & - & - & - & 49.3 & - \\
        UniCLIP-ViT-B/32 & Open30M & 69.2 & 87.8 & 56.5 & 61.1 & 64.6 & 8.0 & 19.5 & 84.0 & 4.7 & 36.6 & \textbf{54.2} & 49.7\\
        \midrule
        \small\textit{Linear probing:}\\
        CLIP-ViT-B/32 & YFCC15M & 71.2 & 89.2 & 72.1 & 70.1 & 71.4 & 93.2 & 34.9 & 84.3 & 29.7 & 60.9 & 61.1 & 67.1 \\
        SLIP-ViT-B/32 & YFCC15M  & 75.4 & 90.5 & 75.3 & 73.5 & 77.1 & 96.1 & 43.0 & 87.2 & 34.1 & 71.1 & 68.1 & 71.9 \\
        DeCLIP-ViT-B/32 & YFCC15M  & 76.5 & 88.6 & 71.6 & 75.9 & 79.3 & 96.7 & 42.6 & 88.0 & 32.6 & 69.1& 69.2 & 71.8\\
        UniCLIP-ViT-B/32 & YFCC15M  & \textbf{83.1} & \textbf{92.5} & \textbf{78.2} & \textbf{77.0} & \textbf{81.3} & \textbf{97.1} & \textbf{49.8} & \textbf{88.9} & \textbf{36.2} & \textbf{72.8} & \textbf{70.8} & \textbf{75.2} \\
        \cdashlinelr{1-14}
        UniCLIP-ViT-B/32 & Open30M & 85.4 & 95.1 & 81.5 & 79.2 & 84.4 & 97.3 & 67.3 & 91.1 & 39.0 & 77.2 & 74.0 & 79.1\\
        \bottomrule
  \end{tabular}
  \end{adjustbox}
  \vspace{-1em}
\end{table}

\paragraph{Datasets}
For reproducibility, we use publicly available datasets for training and evaluation in our experiments, including CC3M~\cite{sharma2018conceptual}, CC12M~\cite{changpinyo2021conceptual}, DeCLIP YFCC15M~\cite{li2022supervision,thomee2016yfcc100m} for training and Pets~\cite{parkhi2012cats}, CIFAR-10, CIFAR-100~\cite{krizhevsky2009learning}, SUN397~\cite{xiao2016sun}, Food-101~\cite{bossard2014food}, Flowers~\cite{nilsback2008automated}, Cars~\cite{krause20133d}, Caltech-101~\cite{fei2004learning}, Aircraft~\cite{maji2013fine}, DTD~\cite{cimpoi2014describing}, ImageNet-1k~\cite{russakovsky2015imagenet}, Flickr30k~\cite{plummer2015flickr30k}, COCO Captions~\cite{chen2015microsoft} for evaluation.
We define the union of CC3M, CC12M, and YFCC15M as Open30M dataset.

\paragraph{Settings}
For each original image and corresponding text caption, one weakly augmented image, two strongly augmented images, and one text form a positive group in our experiments.
Detailed augmentation and optimization configurations can be found in Appendix.

\subsection{Main Results}
\label{main}
We evaluate the transferability of our model in single-modal and multi-modal downstream tasks.
Linear probing and fine-tuning on image classification tasks are performed for single-modal benchmarks, and image--text retrieval tasks and zero-shot image classification tasks are evaluated for multi-modal benchmarks.

\paragraph{Linear Probing \& Fine-Tuning}
\begin{wraptable}{r}{0.45\textwidth}
    \vspace{-1.3em}
    \caption{ImageNet-1k fine-tuning accuracy for the models pre-trained on YFCC15M.}
    \label{exp:fine}
    \centering
    \begin{tabular}{cc}
        \toprule
        Method & Accuracy \\
        \midrule
        CLIP-ViT-B/32 & 72.27 \\
        SLIP-ViT-B/32 & 75.64 \\
        DeCLIP-ViT-B/32 & 74.34\\
        UniCLIP-ViT-B/32 & \textbf{76.54} \\
        \bottomrule
    \end{tabular}
    \vspace{-2.5em}
\end{wraptable}
For single-modal experiments, we remove the image projection head $g_I$ and augmentation encoder $f_A$, and use only the image encoder $f_I$. Table \ref{exp:main} reports linear classification performances on 11 downstream datasets.
We report ImageNet fine-tuning accuracy in Table \ref{exp:fine}. UniCLIP consistently outperforms other methods on all downstream datasets in the single-modal experiments.

\paragraph{Zero-Shot Classification \& Image--Text Retrieval}
Table \ref{exp:main} shows zero-shot classification performances on 11 downstream datasets. We perform prompt ensembling for each class with the same prompt templates as \cite{mu2021slip, radford2021learning}.
Table \ref{result:retrieval} shows the results of zero-shot image--text retrieval on Flickr30k and COCO Captions benchmarks.
\begin{table}[t]
    \caption{Zero-shot image--text retrieval on the test splits of Flickr30k and COCO Captions with models pre-trained on YFCC15M. $^\dagger$Pre-trained on Open30M.}
    \label{result:retrieval}
    \centering
        \begin{adjustbox}{width=1.0\columnwidth,center}
        \centering
        \begin{tabular}{lccc|ccc|ccc|ccc}
        \toprule
        & \multicolumn{6}{c}{Image-to-text retrieval} & \multicolumn{6}{c}{Text-to-image retrieval} \\
        & \multicolumn{3}{c}{Flickr30k} & \multicolumn{3}{c}{COCO Captions} & \multicolumn{3}{c}{Flickr30k} & \multicolumn{3}{c}{COCO Captions} \\
        Method &  R@1 & R@5 & R@10 & R@1 & R@5 & R@10 & R@1 & R@5 & R@10 & R@1 & R@5 & R@10 \\
        \midrule
        CLIP-ViT-B/32 & 34.9	& 63.9 & 75.9 & 20.8 & 43.9 & 55.7 & 23.4 & 47.2 & 58.9 & 13.0 & 31.7 & 42.7\\
        SLIP-ViT-B/32 & 47.8	& 76.5 & 85.9 & 27.7 & 52.6 & 63.9 & 32.3 & 58.7 & 68.8 & 18.2 & 39.2 & 51.0\\
        DeCLIP-ViT-B/32 & 51.4 & 80.2 & 88.9 & 28.3 & 53.2 & 64.5 & 34.3 & 60.3 & 70.7 & 18.4 & 39.6 & 51.4\\
        UniCLIP-ViT-B/32 & \textbf{52.3} & \textbf{81.6} & \textbf{89.0} & \textbf{32.0} & \textbf{57.7} & \textbf{69.2} & \textbf{34.8} & \textbf{62.0} & \textbf{72.0} & \textbf{20.2} & \textbf{43.2} & \textbf{54.4}\\
        \cdashlinelr{1-13}
        UniCLIP-ViT-B/32$^\dagger$ & 75.6 & 94.2 & 97.3 & 46.1 & 74.0 & 83.0 & 61.4 & 85.2 & 91.5 & 35.2 & 61.3 & 71.7\\
        \bottomrule
        \end{tabular}
        \end{adjustbox}
\end{table}

\subsection{Ablation Studies}
\label{abl}
In this section, we report experiments to inspect how each component of UniCLIP contributes to the final performance.
We pre-train all variants of UniCLIP with a ViT-B/16 backbone on the CC3M dataset for 50 epochs, and compare their ImageNet-1k zero-shot evaluation performances.

\paragraph{Image Projection Head Types}
We tried several different architectures for the image projection head including a linear layer, MLP layers, and residual blocks, when the head takes augmentation embeddings as input or not, as in Table~\ref{exp:head}.
It turns out that MLP shows a strong tendency to overfit, even performs worse than a linear layer.
By adding skip connections to the head, it can fully utilize augmentation information with increased capabilities while avoiding overfitting.
Making the image projection head augmentation-aware improves the performance for all head types as the head can handle the inter-domain misalignments.

\paragraph{Augmentation Configurations}
Table~\ref{exp:augtoken} studies the effect of augmentation encoding on various augmentation configurations.
Using only strongly augmented images severely degrades the performance without augmentation embedding, because strong augmentations will generate more image--text misalignments.
Since it is observed that including one weakly augmented image works better than using only strongly augmented ones, we choose to keep one weakly augmented image in the positive set as a stable reference sample.

\begin{table}[h]
    \caption{ImageNet-1k zero-shot accuracy with varying image projection head types and augmentation configurations.}
    \begin{subtable}[t]{0.45\textwidth}
        \centering
        \caption{\textbf{Image projection head types.} One weak and two strong image augmentations are used.}
        \begin{adjustbox}{max width=\linewidth}
        \begin{tabular}{ccc}
        \toprule
        \shortstack{Augmentation \\ embedding} & Head type & Accuracy \\
        \midrule
        \multirow{4}{*}{\xmark} &  MLP 3 layers & 24.01 \\ 
        & MLP 6 layers & 23.62\\
        & 1 ResBlock & \textbf{24.76}\\
        & 3 ResBlocks & 24.46\\
        \midrule
        \multirow{5}{*}{\cmark} & Linear layer & 24.68 \\
        & MLP 3 layers & 24.54\\
        & MLP 6 layers & 24.15 \\
        & 1 ResBlock & 27.67 \\
        & 3 ResBlocks & \textbf{27.84} \\
        \bottomrule
        \end{tabular}
        \end{adjustbox}
        \label{exp:head}
    \end{subtable}
    \hfill
    \begin{subtable}[t]{0.45\textwidth}
        \centering
        \caption{\textbf{Augmentation configurations.} 1-ResBlock head is used for no augmentation embedding config and 3-ResBlock head is used with augmentation embedding.  }
        \begin{adjustbox}{max width=\linewidth}
        \begin{tabular}{ccc}
        \toprule
        \shortstack{Augmentation \\ embedding} & Augmentation & Accuracy \\
        \midrule
        \multirow{3}{*}{\xmark} &  3 weak & 24.49 \\ 
        & 1 weak, 2 strong & \textbf{24.76}\\
        &  3 strong & 22.60\\
        \midrule
        \multirow{3}{*}{\cmark} & 3 weak &  23.40\\
        &1 weak, 2 strong & \textbf{27.84}\\
        & 3 strong & 26.43\\
        \bottomrule
        \end{tabular}
        \end{adjustbox}
        \label{exp:augtoken}
    \end{subtable}
    \vspace{1em}
\end{table}

\paragraph{Domain-dependent Similarity Score and Unified Supervision}
In Table~\ref{abl:unitemp}, we can see how the performance changes depending on whether the shared similarity score (Equation~\ref{eq:sim_th}) or the domain-dependent score (Equation~\ref{eq:newsim}) is used.
We also run experiments where positive and negative sets are formed separately with respect to the domain combination as in SLIP~\cite{mu2021slip} and DeCLIP~\cite{li2022supervision}.
The best performance comes out from the domain-dependent similarity measure with unified supervisions, as expected.

\begin{table}[h]
    \centering
    \caption{ImageNet-1k zero-shot accuracy with domain-dependency of similarity score and supervision.}
    \label{abl:unitemp}
    \begin{tabular}{ccc}
        \toprule
        Temperature and offset & Supervision & Accuracy \\
        \midrule
        Shared across domains & Unified & 25.51\\
        Domain-dependent & Separated & 26.59\\
        Domain-dependent & Unified & \textbf{27.84}\\
        \bottomrule
    \end{tabular}
\end{table}

\paragraph{Loss Functions}

As analyzed in Section~\ref{method:loss}, SupCon loss~\cite{khosla2020supervised} outperforms MIL-NCE loss~\cite{miech2020end}, but performs worse than the multi-positive version of InfoNCE loss (Equation~\ref{eq:vanila}), as in Table~\ref{exp:loss}.
The balancing weight $w_{\mathcal{D}(i,p)}$ can boost the performance, and surprisingly, we can significantly improve performance with negligible additional computations by simply adding a trivial pair $(z_i, z_i)$ to the positive set $P_i$.

\begin{table}[h]
\centering
    \caption{ImageNet-1k zero-shot accuracy with different loss functions.}
    \begin{tabular}{cc}
    \toprule
    \centering
    Loss function &  Accuracy\\
    \midrule
    MIL-NCE & 22.23 \\
    SupCon & 23.04 \\
    MP-NCE w/o trivial pair $(z_i, z_i)$ and $w_{\mathcal{D}(i,p)}$ (Eq. \ref{eq:vanila}) & 24.60 \\
    MP-NCE w/o $w_{\mathcal{D}(i,p)}$ (Eq. \ref{eq:ours_noweight}) & 26.41 \\
    MP-NCE & \textbf{27.84} \\
    \bottomrule
    \end{tabular}
    \label{exp:loss}
\end{table}

\section{Conclusion}
We have proposed UniCLIP, a unified framework for visual--language pre-training that improves data-efficiency by integrating contrastive losses defined across multiple domains into a single universal space.
In this paper, image--text datasets were used to validate our method since vision and language are among the most actively studied fields in deep learning. Although we have experimented with vision--language multimodal datasets only, the proposed UniCLIP framework can be easily extended to other types of multimodal datasets because it is designed in a modality-agnostic way except for the augmentation encoding part. All modality-specific knowledge required to apply UniCLIP to different types of modality is to describe each modality-specific augmentation as a real vector, as in Section~\ref{method:arch}, which is quite straightforward.
We leave it for future work to see how well the UniCLIP framework works with various types of multimodal datasets.

\section*{Acknowledgements}
This work was supported by Institute of Information \& communications Technology Planning \& Evaluation (IITP) grant funded by the Korea government(MSIT). (No. 2022-0-00184, Development and Study of AI Technologies to Inexpensively Conform to Evolving Policy on Ethics)

% \newpage
{\small
\bibliographystyle{abbrvnat}
\bibliography{egbib}
}

\newpage
\input{appendix}

\end{document}

%% file: abstract.tex
\begin{abstract}
Pre-training vision--language models with contrastive objectives has shown promising results that are both scalable to large uncurated datasets and transferable to many downstream applications.
Some following works have targeted to improve data efficiency by adding self-supervision terms, but inter-domain (image--text) contrastive loss and intra-domain (image--image) contrastive loss are defined on individual spaces in those works, so many feasible combinations of supervision are overlooked.
To overcome this issue, we propose UniCLIP, a Unified framework for Contrastive Language--Image Pre-training.
UniCLIP integrates the contrastive loss of both inter-domain pairs and intra-domain pairs into a single universal space.
The discrepancies that occur when integrating contrastive loss between different domains are resolved by the three key components of UniCLIP: (1) augmentation-aware feature embedding, (2) MP-NCE loss, and (3) domain dependent similarity measure.
UniCLIP outperforms previous vision--language pre-training methods on various single- and multi-modality downstream tasks.
In our experiments, we show that each component that comprises UniCLIP contributes well to the final performance.
\end{abstract}

%% file: introduction.tex
\section{Introduction}
Recent advances in deep learning have shown significant progress in pre-training large-scale models that transfer well to various downstream applications.
Following the success of this paradigm in both fields of computer vision and natural language processing, vision--language pre-training models~\cite{jia2021scaling,radford2021learning} that learn image representations from natural language supervision have been proposed.
In those works, pre-training is done under a simple contrastive loss that makes the embedding of an image and its matching text description (positive pair) more similar to each other than other arbitrary image--text pairs (negative pairs).

Towards a more data-efficient pre-training objective, subsequent works~\cite{li2022supervision,mu2021slip} introduced additional self-supervision terms to the image--text contrastive loss, including self-supervision for augmented images~\cite{chen2020simple, chen2021exploring}, augmented texts~\cite{wei2019eda}, and masked texts~\cite{li2022supervision}.
Involving more pairs of positive/negative supervisions into the final contrastive loss leads to a more mathematically pleasing objective~\cite{chen2020simple}, thus enabling the model to be more data-efficient.
Yet, these works entail a major limitation since the contrastive loss for intra-domain pairs, such as image--image pairs, and inter-domain pairs, such as image--text pairs, are defined independently in separated spaces.
This means that the contrastive loss is unaware of a substantial set of feasible combinations for negative supervision, for instance image--image pairs are not included when calculating the contrastive loss for image--text supervision, leaving a huge room for improvement in terms of data-efficiency and feature-diversity.
Based on this observation, we set the goal of this paper to build a contrastive image--text pre-training framework where the contrastive learning of all possible intra-domain and inter-domain pairs is defined in the same single unified embedding space.

\begin{wrapfigure}{r}{0.4\textwidth}
    \vspace{-1em}
    \begin{center}
        \includegraphics[width=0.4\textwidth]{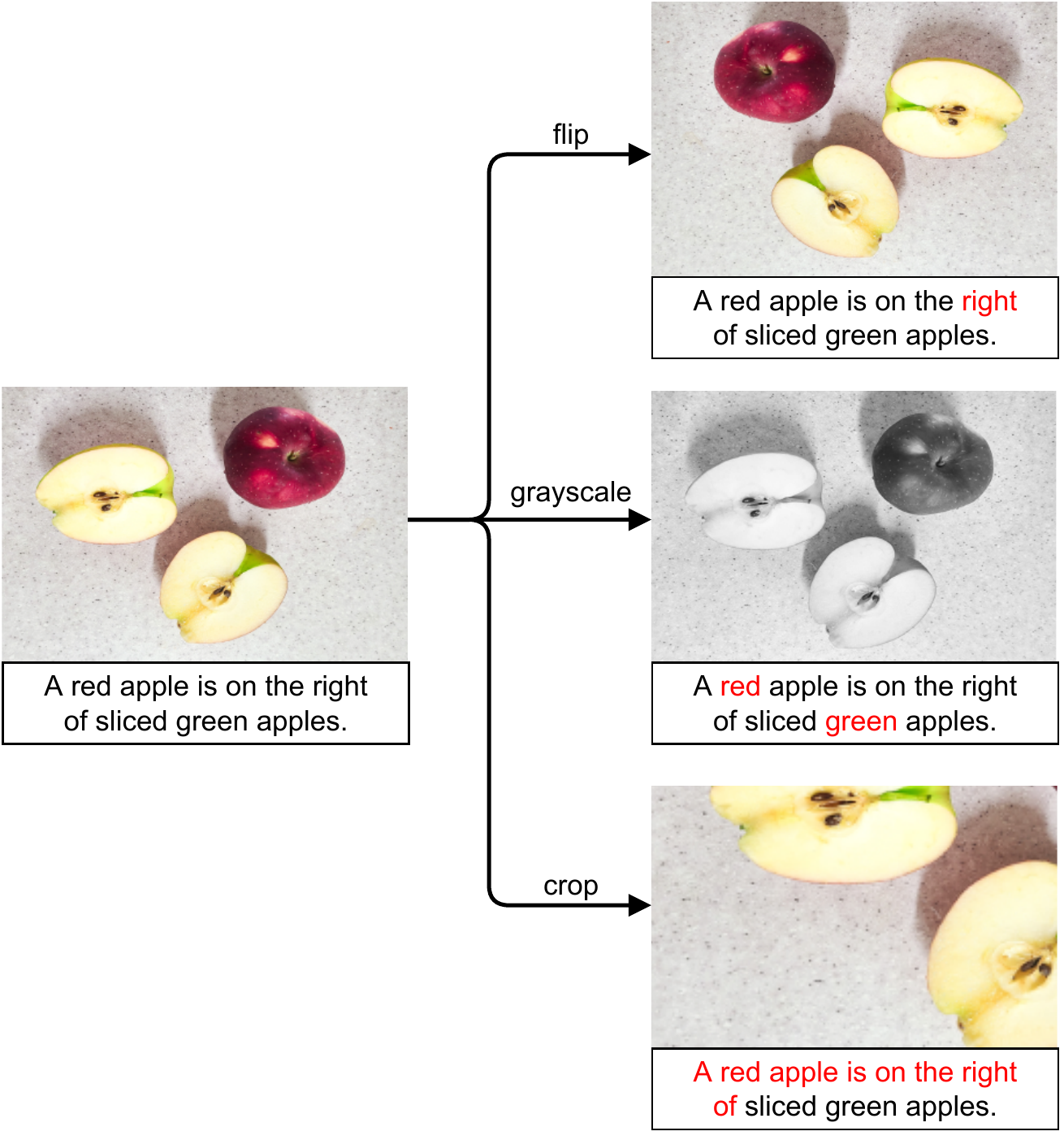}
        \caption{Image--text misalignments caused by data augmentations. The misaligned texts are highlighted in {\color{red}red} (best viewed in color).}
        \label{fig:misalign}
    \end{center}
    \vspace{-1em}
\end{wrapfigure}

Though this goal sounds intuitive, defining a contrastive loss between the multiple modalities in a unified space has several challenges.
First, misalignments can occur between the image--text semantics when applying image augmentations.
For example in Figure~\ref{fig:misalign}, the semantic of `a red apple is on the right of sliced green apples' can be easily broken by simple image augmentations like horizontal flipping, converting to grayscale, or cropping, whereas they are fundamental augmentations used in image--image contrastive self-supervised learning~\cite{chen2020simple}.
We validate from our experiments that leaving this discrepancy unattended hinders training and degrades final performance.
Secondly, existing contrastive losses in literature for multi-positive pairs~\cite{khosla2020supervised,miech2020end} are not compatible with our training objective that deals with embedding of different modalities.
This is because intra-domain pairs, like two different augmented views of a single image, serve as relatively easier examples than inter-domain pairs like image--text pairs.
Existing losses~\cite{khosla2020supervised,miech2020end} are vulnerable to this condition as easy-positive examples and hard-positive examples interfere with each other.
Lastly, we discovered that applying the same similarity measure between embeddings from different modalities in our contrastive loss results in a suboptimal performance,
because there are inherent differences in similarity measures between inter-domain and intra-domain pairs, \textit{i.e.}, samples in an intra-domain pair can be arbitrarily close but samples in an inter-domain pair cannot.

In this paper, we propose UniCLIP: a \textbf{Uni}fied framework for \textbf{C}ontrastive \textbf{L}anguage--\textbf{I}mage \textbf{P}re-training, that unifies contrastive objectives between multiple modalities on a single embedding space.
Each challenge above is addressed with our key components of UniCLIP: (1) \textbf{augmentation-aware feature embedding} that makes UniCLIP aware of misalignments caused by data augmentations, (2) \textbf{MP-NCE loss} that is designed to stabilize training for both easy- and hard-positive pairs, and (3) \textbf{domain dependent similarity measure} that adjusts the difference in similarity scales between inter-domain pairs and intra-domain pairs.
UniCLIP outperforms existing vision--language pre-training methods in various single- and multi-modal downstream tasks such as linear probing, zero-shot classification, fine-tuning, and image--text retrieval, by addressing the three problems described above.
We validate that each component of UniCLIP successfully addresses the issues of contrastive learning in a unified space and meaningfully contributes to the final performance.
Our contribution is summarized as follows:
\begin{itemize}
    \item We propose UniCLIP, a unified framework for visual--language pre-training that improves data-efficiency by integrating contrastive losses defined across multiple domains into a single universal space.
    We study new technical challenges that occur from this integration.
    
    \item We design new components for UniCLIP to address the aforementioned challenges: augmentation-aware feature embedding, MP-NCE loss, and domain dependent similarity measure. Our extensive experiments show that each of our proposed components serves a key role in the final performance.
    
    \item UniCLIP outperforms existing vision--language pre-training methods across multiple downstream tasks that include various modalities. 
\end{itemize}

%% file: appendix.tex
\appendix

\setcounter{table}{0}
\renewcommand{\thetable}{\Alph{table}}

\setcounter{figure}{0}
\renewcommand{\thefigure}{\Alph{figure}}

\setcounter{algorithm}{0}
\renewcommand{\thealgorithm}{\Alph{algorithm}}

\section{Related Works}
\begin{figure}[h]
    \centering
    \begin{subfigure}[b]{0.3\textwidth}
        \centering
        \includegraphics[width=\textwidth]{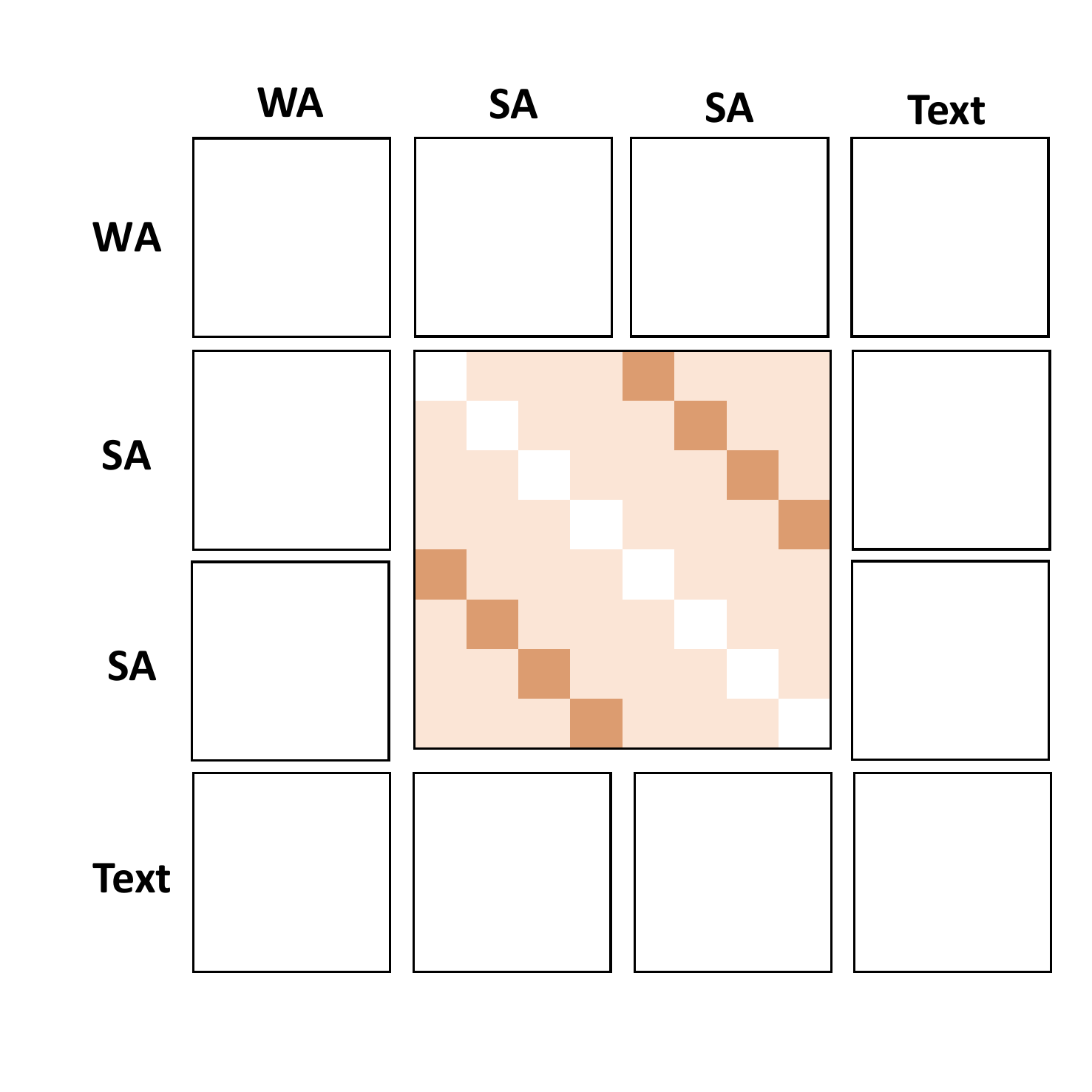}
        \caption{SSL (SimCLR)}
        \label{fig:ssl}
    \end{subfigure}
    \hfill
    \begin{subfigure}[b]{0.3\textwidth}
        \centering
        \includegraphics[width=\textwidth]{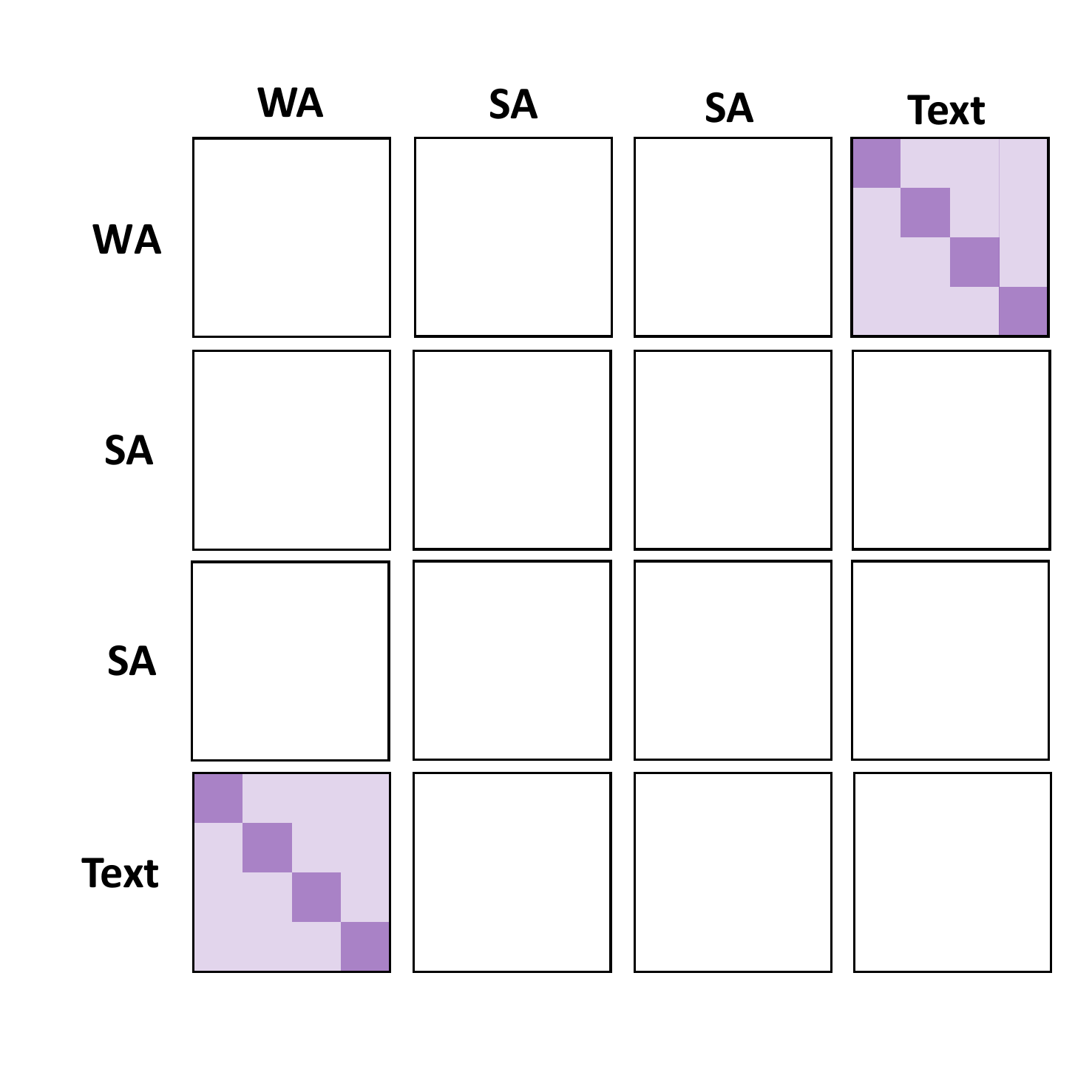}
        \caption{CLIP}
        \label{fig:clip}
    \end{subfigure}
    \hfill
    \begin{subfigure}[b]{0.3\textwidth}
        \centering
        \includegraphics[width=\textwidth]{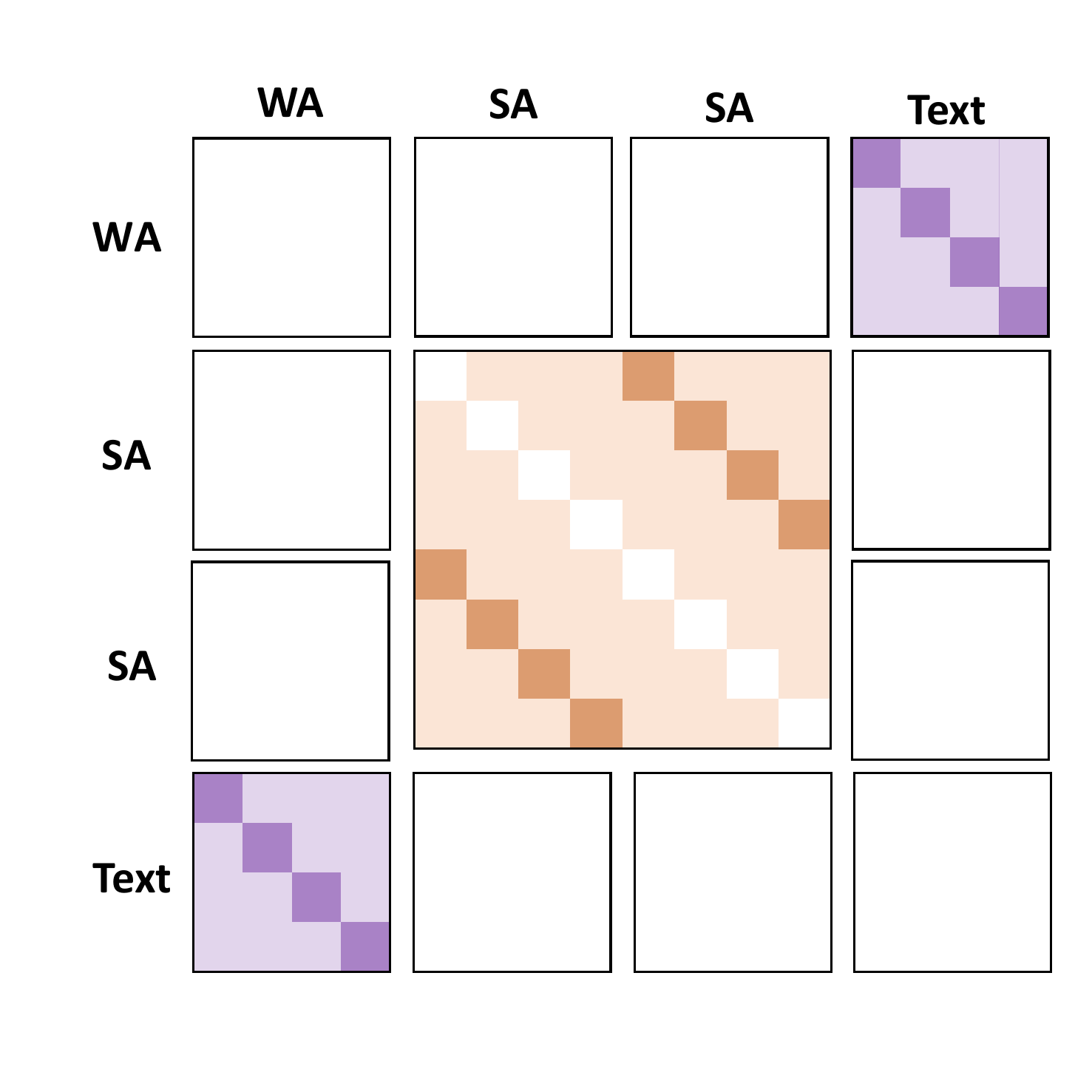}
        \caption{SLIP}
        \label{fig:slip}
    \end{subfigure}
    \hfill
    \begin{subfigure}[b]{0.3\textwidth}
        \centering
        \includegraphics[width=\textwidth]{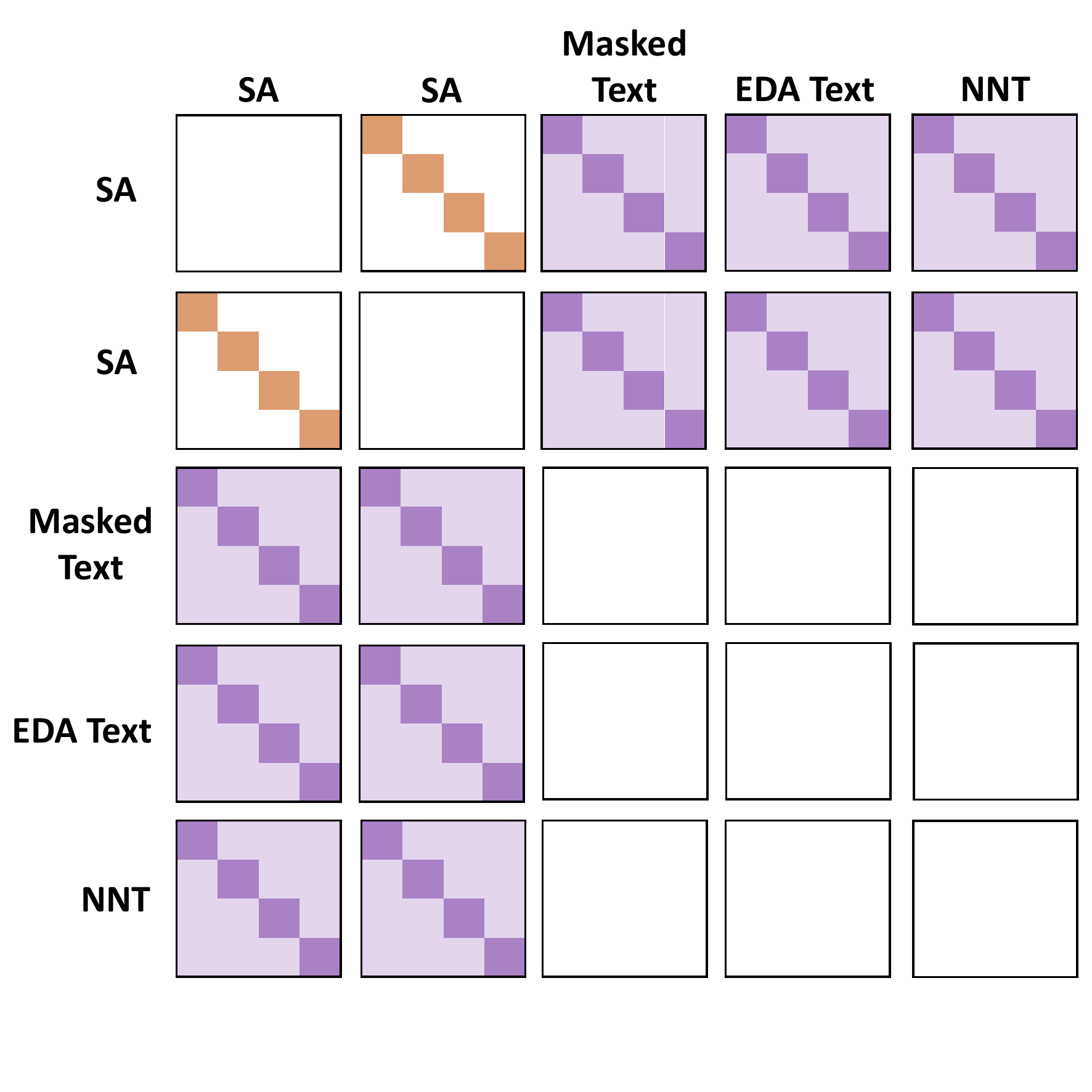}
        \caption{DeCLIP}
        \label{fig:declip}
    \end{subfigure}
    \hfill
    \begin{subfigure}[b]{0.3\textwidth}
        \centering
        \includegraphics[width=\textwidth]{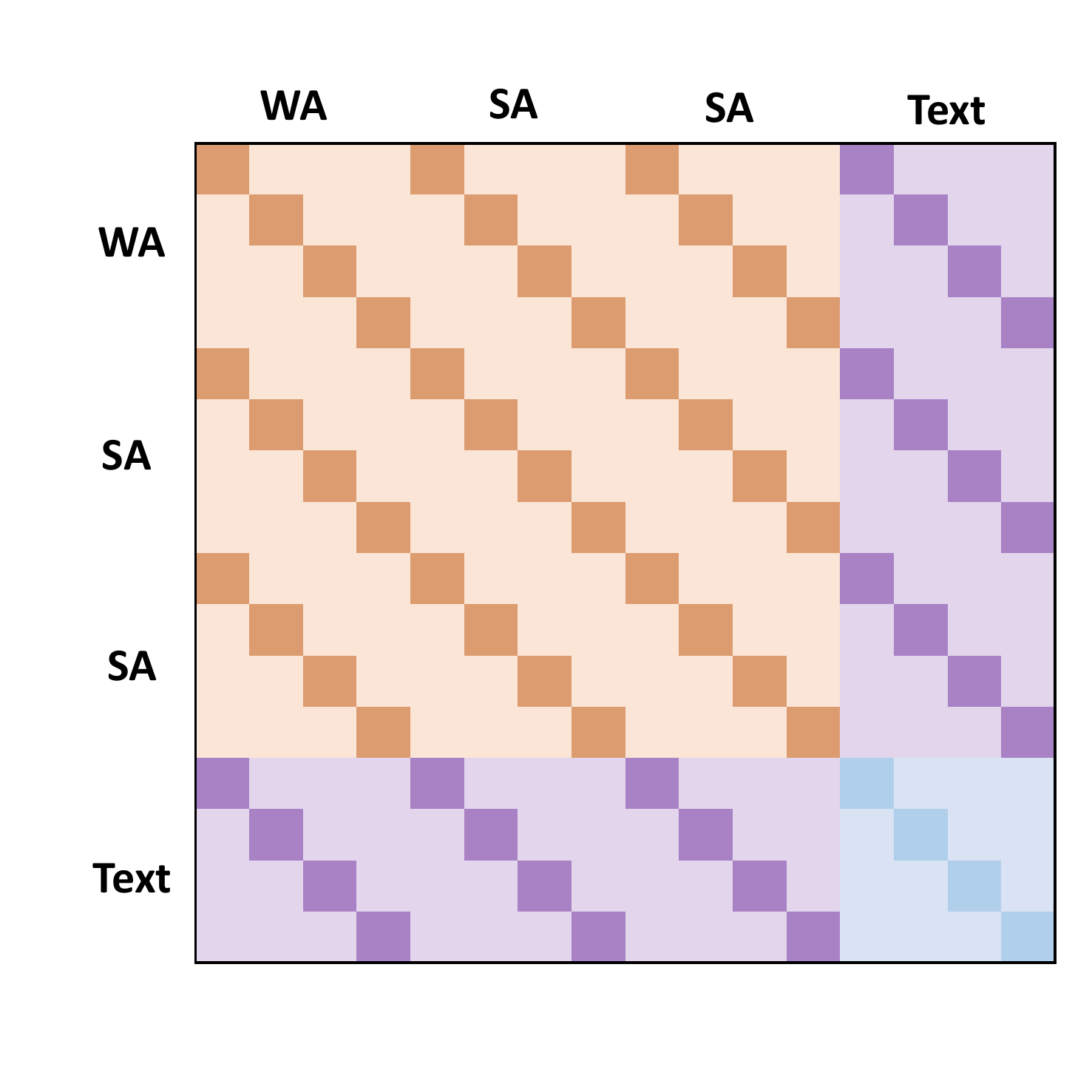}
        \caption{UniCLIP -- Ours}
        \label{fig:ours}
    \end{subfigure}
    \hfill
    \begin{subfigure}[b]{0.3\textwidth}
        \centering
        \includegraphics[width=\textwidth]{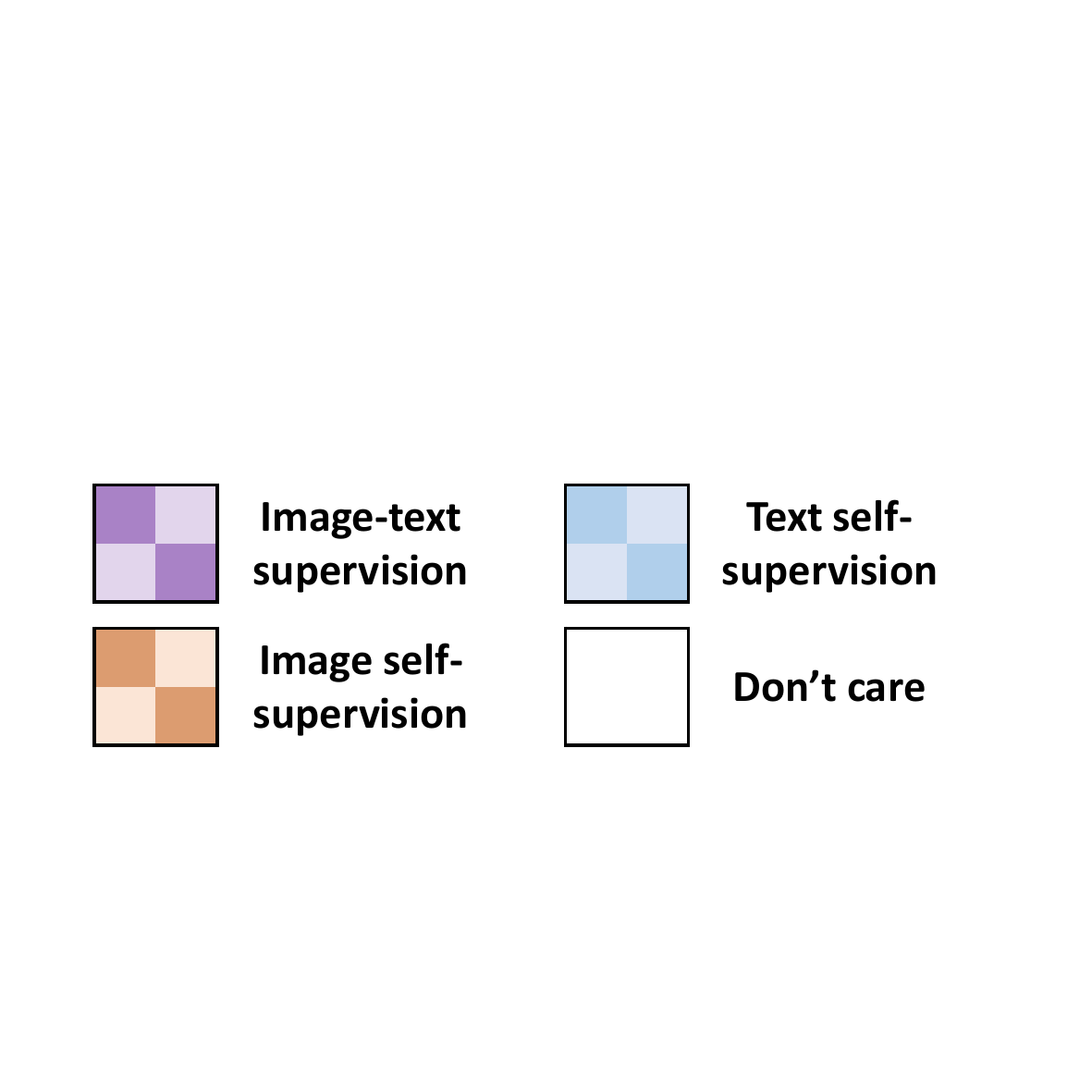}
    \end{subfigure}
    \caption{Similarity matrices in various contrastive learning methods. Darker colors represent positive pairs and lighter colors represent negative pairs. \textbf{WA}: Weakly Augmented image, \textbf{SA}: Strongly Augmented image, \textbf{NNT}: Nearest Neighborhood Text.}
    \label{fig:related}
\end{figure}

\begin{figure}[h]
    \centering
    \begin{adjustbox}{width=1.0\textwidth,center}
    \includegraphics[]{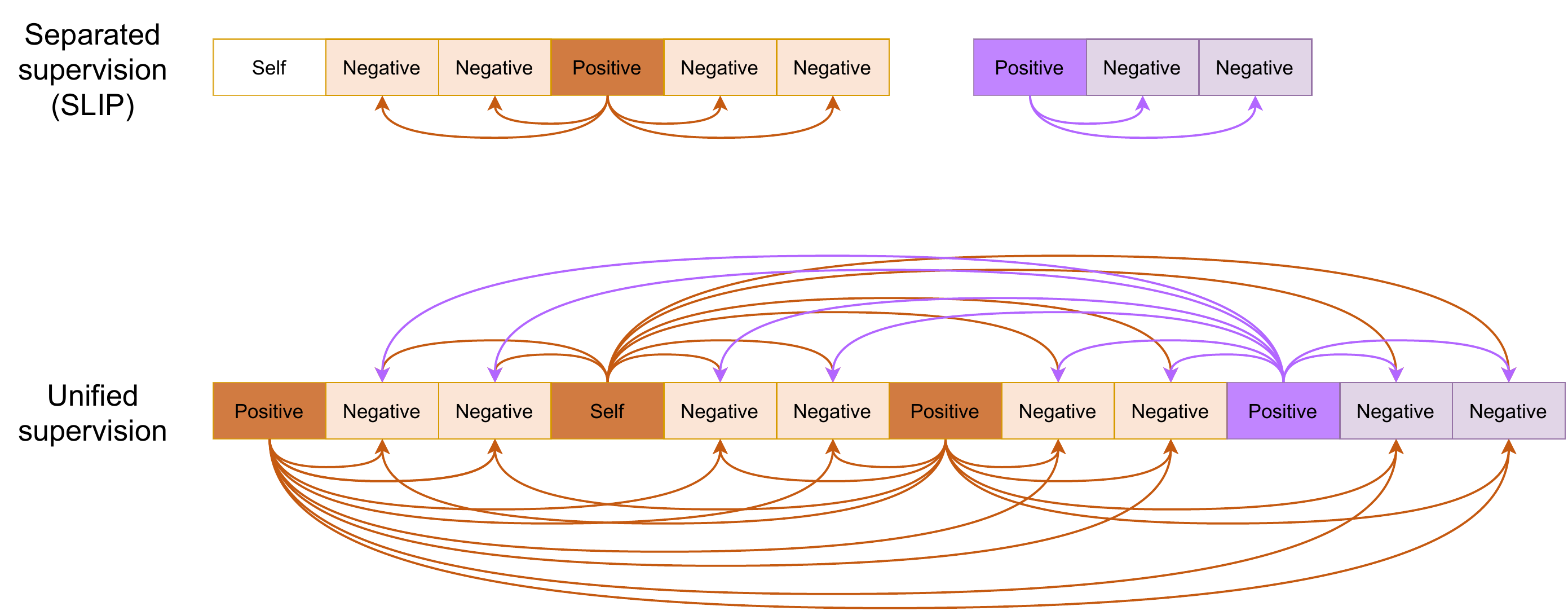}
    \end{adjustbox}
    \caption{Separated supervision and unified supervision. All possible pairs including intra-domain and inter-domain pairs contribute to contrastive learning across different supervisions in our unified framework, whereas each supervision is considered independently in previous works.}
    \label{fig:sepuni}
\end{figure}

\paragraph{Self-Supervised Learning}

Recently, self-supervised learning (SSL) has drawn a huge attention as a pre-training method that is scalable to large and uncurated datasets.
Among the various pretext tasks proposed for self-supervised learning~\cite{doersch2015unsupervised,noroozi2016unsupervised,zhang2016colorful,pathak2016context,zhang2017split,noroozi2017representation,doersch2017multi,gidaris2018unsupervised}, it has been demonstrated that minimizing the contrastive cross entropy loss between positive pairs (\textit{i.e.}, an augmented view of the original data) against negative pairs (\textit{i.e.}, other data samples) yields representations that show solid performance throughout multiple tasks and datasets.

\paragraph{Contrastive Language--Image Pre-training}
CLIP~\cite{radford2021learning} introduced a new paradigm of pre-training by defining a contrastive loss with large-scale image--text pairs (Figure~\ref{fig:clip}).
Here, the image and its matching text description comprise a positive pair, and the representations of the two are learned to be similar to each other than other arbitrary image--text pairs.
CLIP learns powerful representations that are transferable throughout a wide set of datasets and tasks, showing robust performance even in zero-shot evaluations.
The image--text representation obtained by CLIP revolutionized multiple research directions in various fields since it provides a standard measure for how semantically similar a given image--text pair is~\cite{kim2021diffusionclip,ramesh2021zero}.

\paragraph{Self-Supervised Learning Meets CLIP}

Following works of SLIP~\cite{mu2021slip} and DeCLIP~\cite{li2022supervision} improved CLIP by introducing additional SSL terms to the original image--text contrastive loss formula.
However, these methods have limited supervisions since the contrastive loss between inter-domain pairs and intra-domain pairs are defined in separate spaces.
To address this issue, our UniCLIP defines the contrastive loss of both inter-domain pairs and intra-domain pairs in a single unified space, utilizing the supervision from all possible combinations throughout multiple domains at once.

\section{Algorithm}
The main algorithm of UniCLIP is summarized in Algorithm~\ref{alg}.

\begin{algorithm}[h!]
\caption{UniCLIP}
\label{alg}
    \begin{algorithmic}[1]
        \Require image encoder $f_I$, text encoder $f_T$, image projection head $g_I$, text projection head $g_T$, 
        \Statex ~~~~~~augmentation encoder $f_A$, batch size $N$, temperature $\tau \in \mathbb{R}^3$, offset $b \in \mathbb{R}^3$,
        \Statex ~~~~~~weak augmentation distribution $p_{wa}$, strong augmentation distribution $p_{sa}$
        \For{sampled mini-batch $\{(x_k^I,x_k^T)\}_{k=1}^N$} 
            \For{\textbf{all} $k\in \{1, \ldots , N\}$}
            \State draw augmentation instructions $\mathcal{A}_1 \sim p_{wa}$, $\mathcal{A}_2 \sim p_{sa}$, $\mathcal{A}_3 \sim p_{sa}$
            \State $z_{k} = g_I(f_I(\mathcal{A}_1(x_k^I)), f_A(\mathcal{A}_1))$
            \State $z_{k+N} = g_I(f_I(\mathcal{A}_2(x_k^I)), f_A(\mathcal{A}_2))$
            \State $z_{k+2N} = g_I(f_I(\mathcal{A}_3(x_k^I)), f_A(\mathcal{A}_3))$
            \State $z_{k+3N} = g_T(f_T(x_k^T))$
            \EndFor
            \For{\textbf{all} $i \in \{1,\ldots,4N\}$}
            \For{\textbf{all} $j \in \{1,\ldots,4N\}$}
            \State $\mathcal{D}(i,j) = \begin{cases}1,\quad\text{if~}i\le3N\text{~and~}j\le3N\\3,\quad\text{if~}i>3N\text{~and~}j>3N\\2,\quad\text{otherwise}\end{cases}$
            \State $s_{i,j} = \exp \left( \frac{1}{\tau_{\mathcal{D}(i,j)}} \left(\frac{z_i^\top z_j}{\|z_i\| \|z_j\|} - b_{\mathcal{D}(i,j)} \right) \right) $
            \EndFor
            \State $P_i = \{j \in \{1,\ldots,4N\} \setminus \{i\}|(j-i)/N \in \mathbb{Z}\}$
            \State $N_i = \{1,\ldots,4N\} \setminus P_i \setminus \{i\}$
            \State $w = (1/9, 1/6, 1)$
            \State $\mathcal{L}_i = \mathbb{E}_{p \in P_i \cup \{i\}} \left[ - w_{\mathcal{D}(i, p)}  \log \frac{s_{i, p}}{s_{i, p}+\sum_{n \in N_i} s_{i, n}} \right]$
            \EndFor
            \State $\mathcal{L} = \frac{1}{4N}\sum_{i=1}^{4N} \mathcal{L}_i$
            \State update networks, temperature, offset to minimize $\mathcal{L}$
        \EndFor
    \end{algorithmic}
\end{algorithm}

\newpage
\section{Additional Experimental Results}

\paragraph{Zero-Shot Classification on ImageNet Variations}
We report zero-shot classification performance on ImageNet variations such as ImageNet-R~\cite{hendrycks2021many}, ImageNet-Sketch~\cite{wang2019learning}, ImageNetV2~\cite{recht2019imagenet}, and ImageNet-A~\cite{hendrycks2021natural} in Table~\ref{exp:robust}.
\begin{table}[h]
    \caption{Zero-shot accuracy on ImageNet variations.}
    \label{exp:robust}
    \centering
    \begin{adjustbox}{width=1.0\columnwidth,center}
    \begin{tabular}{lcccccc}
        \toprule
        Method & \shortstack{Pre-train \\ dataset} & ImageNet & ImageNet-R & ImageNet-Sketch & ImageNetV2 & ImageNet-A\\
        \midrule
        CLIP-ViT-B32 & YFCC15M & 31.3 & 22.6 & 7.2 & 25.5/30.6/33.6 & 8.1\\
        SLIP-ViT-B32 & YFCC15M & 38.3 & 31.7 & 11.9 & 33.2/37.8/41.8 & 13.2\\
        DeCLIP-ViT-B32 & YFCC15M & 41.2 & 34.3 & 14.5 & 35.4/40.4/43.8 & \textbf{15.0}\\
        UniCLIP-ViT-B32 & YFCC15M & \textbf{42.8} & \textbf{37.8} & \textbf{15.7}& \textbf{36.5/41.9/46.3} & 14.4\\
        \cdashlinelr{1-7}
        UniCLIP-ViT-B/32 & Open30M & 54.2 & 61.8 & 36.0 & 47.1/54.0/58.6 & 18.3\\  
        \bottomrule
    \end{tabular}
    \end{adjustbox}
\end{table}

\paragraph{Zero-Shot Image--Text Retrieval on Validation Splits}
Table \ref{result:retrieval_val} shows the results of zero-shot image--text retrieval on the validation splits of Flickr30k and COCO Captions benchmarks.
\begin{table}[h]
    \caption{Zero-shot image--text retrieval on the validation splits of Flickr30k and COCO Captions with models pre-trained on YFCC15M. $^\dagger$Pre-trained on Open30M.}
    \label{result:retrieval_val}
    \centering
        \begin{adjustbox}{width=1.0\columnwidth,center}
        \centering
        \begin{tabular}{lccc|ccc|ccc|ccc}
        \toprule
        & \multicolumn{6}{c}{Image-to-text retrieval} & \multicolumn{6}{c}{Text-to-image retrieval} \\
        & \multicolumn{3}{c}{Flickr30k} & \multicolumn{3}{c}{COCO Captions} & \multicolumn{3}{c}{Flickr30k} & \multicolumn{3}{c}{COCO Captions} \\
        Method &  R@1 & R@5 & R@10 & R@1 & R@5 & R@10 & R@1 & R@5 & R@10 & R@1 & R@5 & R@10 \\
        \midrule
        CLIP-ViT-B/32 & 37.3 & 66.2 & 77.1 & 20.1 & 42.9 & 55.1 & 24.9 & 49.0 & 60.0 & 13.3 & 31.7 & 42.3\\
        SLIP-ViT-B/32 & 48.7 & 75.2 & 84.7 & 26.9 & 51.9 & 63.8 & 33.1 & 59.0 & 68.8 & 18.2 & 39.6 & 51.1\\
        DeCLIP-ViT-B/32 & 51.3 & 79.3 & 88.7 & 28.1 & 53.6 & 65.2 & 34.8 & 62.2 & 71.5 & 17.9 & 39.8 & 51.6\\
        UniCLIP-ViT-B/32 & \textbf{55.7} & \textbf{82.9} & \textbf{90.0} & \textbf{32.0} & \textbf{58.8} & \textbf{70.3} & \textbf{36.7} & \textbf{62.6} & \textbf{72.4} & \textbf{20.3} & \textbf{43.1} & \textbf{54.5}\\
        \cdashlinelr{1-13}
        UniCLIP-ViT-B/32$^\dagger$ & 76.7 & 94.2 & 96.9 & 47.8 & 74.4 & 84.2 & 62.4 & 86.7 & 92.2 & 35.4 & 61.6 & 72.0\\
        \bottomrule
        \end{tabular}
        \end{adjustbox}
\end{table}
\paragraph{Fine-tuning Results on Image--Text Retrieval}

We fine-tuned YFCC15M pre-trained models on Flickr30k and COCO Captions for 10 epochs and report the results in Table~\ref{result:ft-testretrieval} and Table~\ref{result:ft-valretrieval}.
Our method consistently outperforms on fine-tuned image--text retrieval benchmarks.

\begin{table}[h!]
    \caption{Fine-tuned image--text retrieval on the test splits of Flickr30k and COCO Captions with models pre-trained on YFCC15M. $^\dagger$Pre-trained on Open30M.}
    \label{result:ft-testretrieval}
    \centering
        \begin{adjustbox}{width=1.0\columnwidth,center}
        \centering
        \begin{tabular}{lccc|ccc|ccc|ccc}
        \toprule
        & \multicolumn{6}{c}{Image-to-text retrieval} & \multicolumn{6}{c}{Text-to-image retrieval} \\
        & \multicolumn{3}{c}{Flickr30k} & \multicolumn{3}{c}{COCO Captions} & \multicolumn{3}{c}{Flickr30k} & \multicolumn{3}{c}{COCO Captions} \\
        Method &  R@1 & R@5 & R@10 & R@1 & R@5 & R@10 & R@1 & R@5 & R@10 & R@1 & R@5 & R@10 \\
        \midrule
        CLIP-ViT-B/32 & 57.4 & 84.7 & 90.2 & 34.4 & 63.5 & 75.2 & 40.4 & 69.5 & 79.6 & 24.0 & 50.8 & 63.5 \\
        SLIP-ViT-B/32 & 68.9 & 91.9 & 95.1 & 43.7 & 71.8 & 82.4 & 51.0 & 79.5 & 86.8 & 31.0 & 58.8 & 70.3 \\
        DeCLIP-ViT-B/32 & 73.6 & 93.9 & 97.2 & 47.9 & 75.5 & 84.6 & 55.9 & 83.4 & 90.2 & 33.8 & 62.7 & 74.4 \\
        UniCLIP-ViT-B/32 & \textbf{77.9} & \textbf{95.1} & \textbf{98.0} & \textbf{52.7} & \textbf{78.6} & \textbf{87.4} & \textbf{61.0} & \textbf{85.9} & \textbf{92.2} & \textbf{37.6} & \textbf{66.3} & \textbf{77.0}\\
        \cdashlinelr{1-13}
        UniCLIP-ViT-B/32$^\dagger$ & 87.8 & 98.2 & 99.2 & 62.2 & 85.3 & 91.9 & 70.7 & 91.5 & 95.4 & 45.6 & 73.5 & 82.5 \\
        \bottomrule
        \end{tabular}
        \end{adjustbox}
\end{table}

\begin{table}[h!]
    \caption{Fine-tuned image--text retrieval on the validation splits of Flickr30k and COCO Captions with models pre-trained on YFCC15M. $^\dagger$Pre-trained on Open30M.}
    \label{result:ft-valretrieval}
    \centering
        \begin{adjustbox}{width=1.0\columnwidth,center}
        \centering
        \begin{tabular}{lccc|ccc|ccc|ccc}
        \toprule
        & \multicolumn{6}{c}{Image-to-text retrieval} & \multicolumn{6}{c}{Text-to-image retrieval} \\
        & \multicolumn{3}{c}{Flickr30k} & \multicolumn{3}{c}{COCO Captions} & \multicolumn{3}{c}{Flickr30k} & \multicolumn{3}{c}{COCO Captions} \\
        Method &  R@1 & R@5 & R@10 & R@1 & R@5 & R@10 & R@1 & R@5 & R@10 & R@1 & R@5 & R@10 \\
        \midrule
        CLIP-ViT-B/32 & 58.3 & 84.8 & 91.5 & 36.1 & 65.0 & 76.4 & 43.1 & 71.1 & 80.3 & 24.9 & 51.7 & 64.1 \\
        SLIP-ViT-B/32 & 69.6 & 90.4 & 95.7 & 45.0 & 74.0 & 83.0 & 52.1 & 79.4 & 86.9 & 31.6 & 59.5 & 71.3 \\
        DeCLIP-ViT-B/32 & 75.6 & 93.0 & 96.6 & 48.7 & 77.3 & 86.2 & 57.8 & 83.3 & 90.3 & 34.2 & 63.1 & 74.6 \\
        UniCLIP-ViT-B/32 & \textbf{78.1} & \textbf{94.9} & \textbf{97.7} & \textbf{54.5} & \textbf{80.9} & \textbf{89.1} & \textbf{61.0} & \textbf{86.0} & \textbf{91.9} & \textbf{38.0} & \textbf{67.2} & \textbf{78.0}\\
        \cdashlinelr{1-13}
        UniCLIP-ViT-B/32$^\dagger$  & 88.0 & 97.7 & 99.3 & 63.0 & 86.3 & 92.4 & 72.1 & 92.1 & 95.9 & 46.2 & 73.7 & 83.1 \\
        \bottomrule
        \end{tabular}
        \end{adjustbox}
\end{table}

\section{Additional Ablation Studies}

\paragraph{Augmentation-Agnostic vs. Augmentation-Aware Image Encoder}
Making the image encoder augmentation-agnostic is a key design idea for better generalization in a unified contrastive learning framework.
To verify this, we feed the image encoder with the augmentation embedding instead of the projection head and performed the same experiment with this augmentation-aware image encoder.
As seen in Table~\ref{exp:loc}, the augmentation-aware encoder performs much worse than the augmentation-agnostic encoder.
\begin{table}[h!]
    \centering
    \caption{ImageNet-1k zero-shot accuracy with respect to augmentation-awareness of image encoder.}
    \label{exp:loc}
    \begin{tabular}{cc}\toprule
        Image encoder & Accuracy \\
        \midrule
        Augmentation-aware & 23.25\\
        Augmentation-agnostic & \textbf{27.84}\\ 
        \bottomrule
    \end{tabular}
\end{table}

\paragraph{Number of Augmentations}
We investigate the effect of the number of image views and text views that make up our multi-view batch for a given original image--text pair.
As there exists a trade-off between the number of augmentations and the number of original image--text pairs in a batch, the number of image views and text views should be set to appropriate values for a balanced learning of intra-domain features and inter-domain features.
As seen in Table \ref{abl:view}, using more text views is not helpful and actually hurts performance, which means that the benefits from increasing text views do not outweigh the losses due to decreased diversity in data samples, so we instead increase the number of image views.
We have observed that the performance increases until the number of image views is 4, but the improvement was not that significant compared to the increased training time due to the decrease in the number of original pairs in a batch, so we decided to use 3 image views for faster training with acceptable performance.
We leave it as a future work to find effective text augmentation methods in contrastive learning.

\begin{table}[h]
    \caption{ImageNet-1k zero-shot accuracy with varying the number of image views and text views.}
    \label{abl:view}
    \centering
    \begin{tabular}{cccc}
        \toprule
        \# of image views & \# of text views & \# of original pairs & Accuracy \\
        \midrule
        1 & 1 & 192 & 21.80\\
        2 & 1 & 128 & 25.54\\
        2 & 2 & 96  & 24.60\\
        3 & 1 & 96  & \textbf{27.67}\\
        3 & 2 & 72  & 24.57\\
        4 & 1 & 72  & \textbf{28.25} \\
        \bottomrule
    \end{tabular}
\end{table}

\section{Misalignment Adjustments in UniCLIP}

Several image--text pairs that are vulnerable to inter-domain misalignment due to augmentations and their similarity scores are presented in Tables~\ref{qual:color}--\ref{qual:flip}.
In each table, there are two images and two captions, where the top image and the left caption are the original pair, the bottom image is an augmented image, and the right caption is a modified caption to pair with the augmented image.

For example in Table~\ref{qual:color}, the image of yellow flowers is augmented to the image of orange flowers by a \texttt{ColorJitter} augmentation and as a consequence the corresponding caption on the left should be modified to the right one to correct the misalignment.
Since CLIP and SLIP only employ \texttt{RandomResizedCrop} augmentation on image--text pairs, they do not experience severe inter-domain misalignment issues, resulting in higher similarity scores on the correct pairs.
UniCLIP also produces correct results even if it has been trained with strong augmentations, which means the inter-domain misalignment problem is well addressed.
In contrast, DeCLIP suffers from inter-domain misalignments and shows unpredictable results.

Interestingly, the projection head in UniCLIP can adjust inter-domain misalignments when information about the applied augmentation $\mathcal{A}$ is known to it via the augmentation encoder $f_A$.
As in Tables~\ref{qual:color}--\ref{qual:flip}, even for augmented images, UniCLIP can give the original captions higher similarity scores than the modified captions if augmentation information is provided by adjusting misalignment.
For example in Table~\ref{qual:gray}, if the model knows that grayscale augmentation has been applied to the image and we let the model adjust the misalignment due to the augmentation, then it will try to guess what the color of the apples were if the image was an RGB image, which would be red with high probability in this case, thus putting a higher score on the original caption than the modified one.

\begin{table}[h!]
    \cellspacelimit{2pt}
    \caption{Similarity scores between images and captions. Top row: original image, bottom row: jittered image.}
    \centering
    \begin{tabular}{c Sl ccc}
        \toprule
        Image & Method & \textit{``Flowers of \textbf{yellow} color.''} & \textit{``Flowers of \textbf{orange} color.''}\\
        \midrule
        \multirow{4}{*}{\begin{minipage}{0.12\textwidth}
         \includegraphics[width=\linewidth]{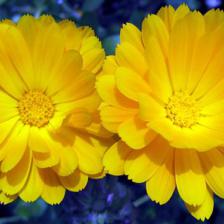}
        \end{minipage}} & CLIP & \textbf{2.6907} & 2.6076 \tabularnewline
         & SLIP & \textbf{2.7534} & 2.7512 \tabularnewline
         & DeCLIP & 0.8485 & \textbf{0.8758} \tabularnewline
         & UniCLIP & \textbf{4.0810} & 2.6024 \tabularnewline
         \midrule
        \multirow{5}{*}{\begin{minipage}{0.12\textwidth}
        \includegraphics[width=\linewidth]{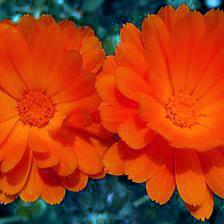}
        \end{minipage}}& CLIP & 2.6213 & \textbf{2.8758} \\
         & SLIP & 2.6075 & \textbf{2.8258} \\
         & DeCLIP & 0.8412 & \textbf{0.8578} \\
         & UniCLIP & 2.8236 & \textbf{3.8955} \\
         & UniCLIP w/ $f_A(\mathcal{A})$ & \textbf{4.0754} & 2.7188 \\
         \bottomrule
    \end{tabular}
    \label{qual:color}
\end{table}

\begin{table}[h!]
    \cellspacelimit{2pt}
    \caption{Similarity scores between images and captions. Top row: original image, bottom row: grayscale image.}
    \centering
    \begin{adjustbox}{width=1.0\columnwidth,center}
    \begin{tabular}{c Sl ccc}
        \toprule
        Image & Method & \textit{``\textbf{Red} apples hanging from the tree.''} & \textit{``\textbf{Gray} apples hanging from the tree.''}\\
        \midrule
        \multirow{4}{*}{\begin{minipage}{0.12\textwidth}
         \includegraphics[width=\linewidth]{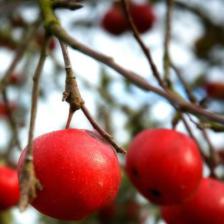}
        \end{minipage}} & CLIP & \textbf{3.3721} & 2.7227 \tabularnewline
         & SLIP & \textbf{3.1031} & 2.4363 \tabularnewline
         & DeCLIP & \textbf{1.3140} & 1.1168 \tabularnewline
         & UniCLIP & \textbf{3.0319} & 2.6291 \tabularnewline
         \midrule
        \multirow{5}{*}{\begin{minipage}{0.12\textwidth}
        \includegraphics[width=\linewidth]{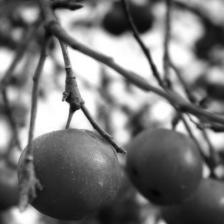}
        \end{minipage}}& CLIP & 3.5971 & \textbf{4.4252} \\
         & SLIP & 2.9678 & \textbf{3.2464} \\
         & DeCLIP & \textbf{1.2837} & 1.1374 \\
         & UniCLIP & 3.5288 & \textbf{3.9235} \\
         & UniCLIP w/ $f_A(\mathcal{A})$ & \textbf{3.0319} & 2.6291 \\
         \bottomrule
    \end{tabular}
    \label{qual:gray}
    \end{adjustbox}
\end{table}

\begin{table}[h!]
    \cellspacelimit{2pt}
    \caption{Similarity scores between images and captions. Top row: original image, bottom row: cropped image.}
    \centering
    \begin{tabular}{c Sl ccc}
        \toprule
        Image & Method & \textit{``A \textbf{tiny} chair.''} & \textit{``A \textbf{close-up} of a chair.''}\\
        \midrule
        \multirow{4}{*}{\begin{minipage}{0.12\textwidth}
         \includegraphics[width=\linewidth]{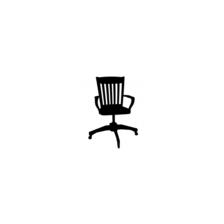}
        \end{minipage}} & CLIP & \textbf{2.1118} & 1.9880 \tabularnewline
         & SLIP & \textbf{2.1535} & 1.8063 \tabularnewline
         & DeCLIP & \textbf{1.1831} & 1.0541 \tabularnewline
         & UniCLIP & \textbf{3.4089} & 2.9047 \tabularnewline
         \midrule
        \multirow{5}{*}{\begin{minipage}{0.12\textwidth}
        \includegraphics[width=\linewidth]{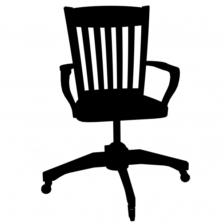}
        \end{minipage}}& CLIP & 2.5919 & \textbf{3.2560} \\
         & SLIP & 2.4766 & \textbf{2.6049} \\
         & DeCLIP & \textbf{1.2042} & 1.0317 \\
         & UniCLIP & 2.5598 & \textbf{2.7660} \\
         & UniCLIP w/ $f_A(\mathcal{A})$ & \textbf{2.1417} & 1.9606 \\
         \bottomrule
    \end{tabular}
    \label{qual:crop}
\end{table}

\begin{table}[h!]
    \cellspacelimit{2pt}
    \caption{Similarity scores between images and captions. Top row: original image, bottom row: flipped image.}
    \centering
    \begin{adjustbox}{width=1.0\columnwidth,center}
    \begin{tabular}{c Sl ccc}
        \toprule
        Image & Method & \textit{``From \textbf{left}, orange, mango, and apple.''} & \textit{``From \textbf{right}, orange, mango, and apple.''}\\
        \midrule
        \multirow{4}{*}{\begin{minipage}{0.12\textwidth}
         \includegraphics[width=\linewidth]{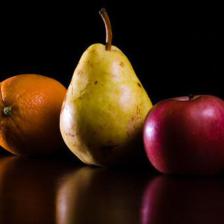}
        \end{minipage}} & CLIP & \textbf{2.6101} & 2.5720 \tabularnewline
         & SLIP & \textbf{2.6285} & 2.5901 \tabularnewline
         & DeCLIP & 1.1800 & \textbf{1.2625} \tabularnewline
         & UniCLIP & \textbf{4.1929} & 3.9725 \tabularnewline
         \midrule
        \multirow{5}{*}{\begin{minipage}{0.12\textwidth}
        \includegraphics[width=\linewidth]{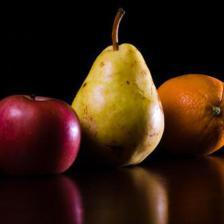}
        \end{minipage}}& CLIP & 2.7728 & \textbf{2.8746} \\
         & SLIP & 2.9061 & \textbf{2.9916} \\
         & DeCLIP & 1.1615 & \textbf{1.2455} \\
         & UniCLIP & 3.9946 & \textbf{4.0283} \\
         & UniCLIP w/ $f_A(\mathcal{A})$ & \textbf{4.2233} & 4.0072 \\
         \bottomrule
    \end{tabular}
    \label{qual:flip}
    \end{adjustbox}
    \vspace{-0.5em}

\end{table}

\section{Analysis}
\begin{figure}[h]
    \centering
    \includegraphics[width=0.45\linewidth]{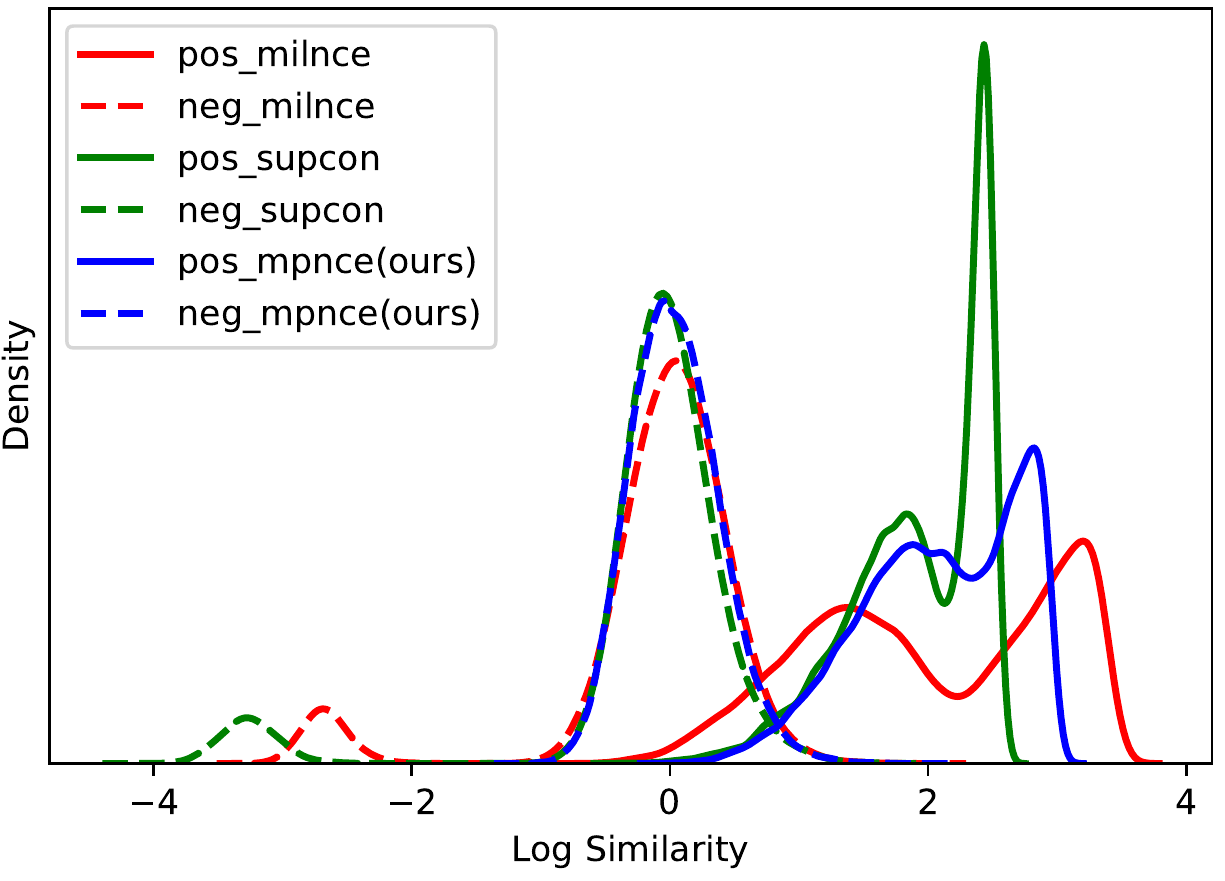}
    \caption{Density plot of similarity scores with respect to loss functions.}
    \label{fig:loss_ftn}
    \vspace{-1.5em}
\end{figure}

\paragraph{Distribution of Similarities w.r.t Loss Functions}
Figure~\ref{fig:loss_ftn} shows distribution of similarities with respect to loss functions. MIL-NCE and SupCon losses show worse separation of positives--negatives compared to MP-NCE loss. In MIL-NCE (red), hard-positives are concentrated around a lower similarity region. SupCon loss (green) shows better separation of positives-negatives, but converges to decreased scores of easy-positives. MP-NCE (blue) shows the best separation of positives--negatives, as well as better convergence of easy-positives compared to SupCon loss. This result is consistent with the analysis in Section~\ref{method:loss}.

\begin{figure}[h]
    \centering
    \begin{subfigure}{0.49\linewidth}
        \centering
        \includegraphics[width=\linewidth]{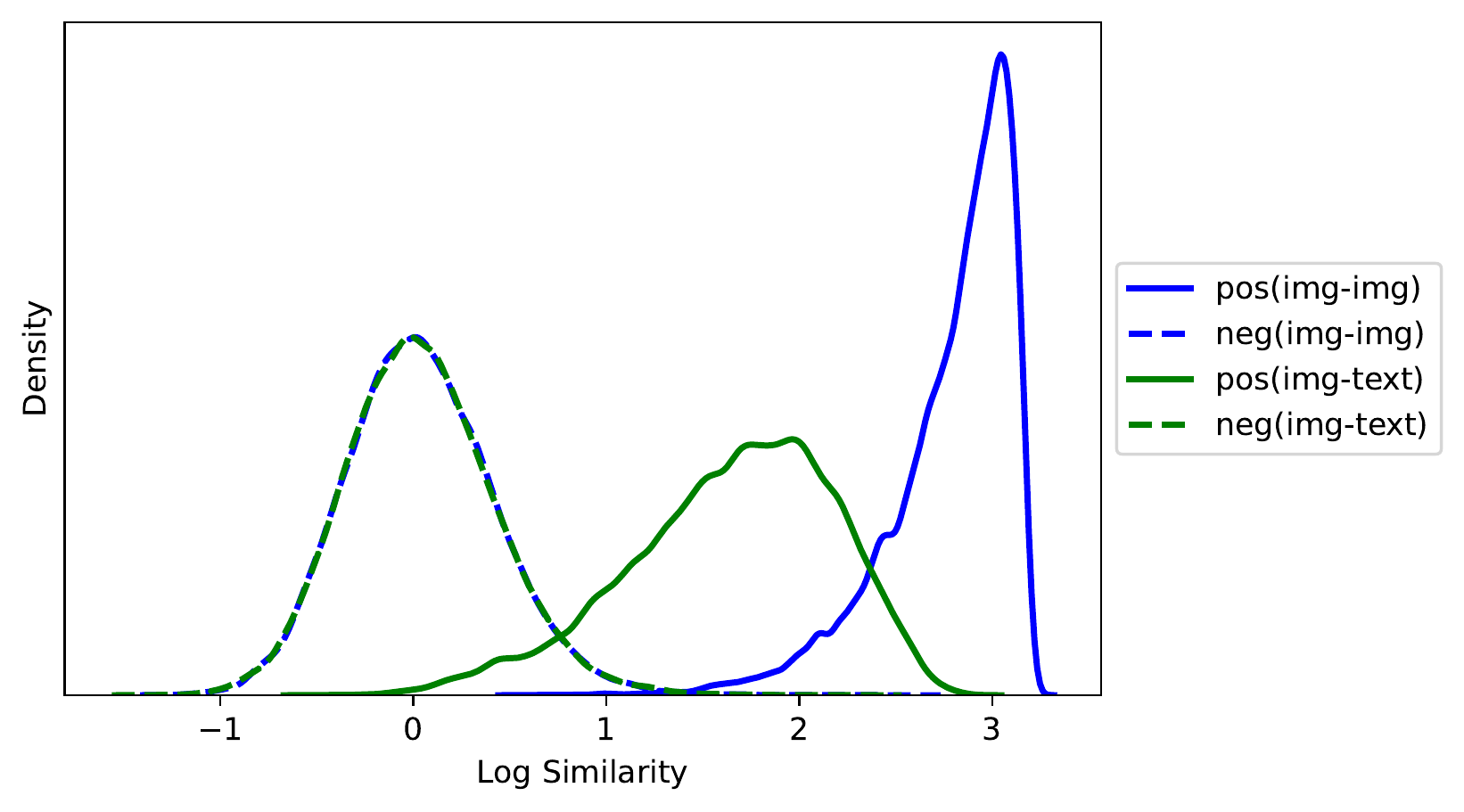}
        \caption{Shared $\tau, b$ across domains.}
        \label{fig:shared}
    \end{subfigure}
    \hfill
    \begin{subfigure}{0.49\linewidth}
        \centering
        \includegraphics[width=\linewidth]{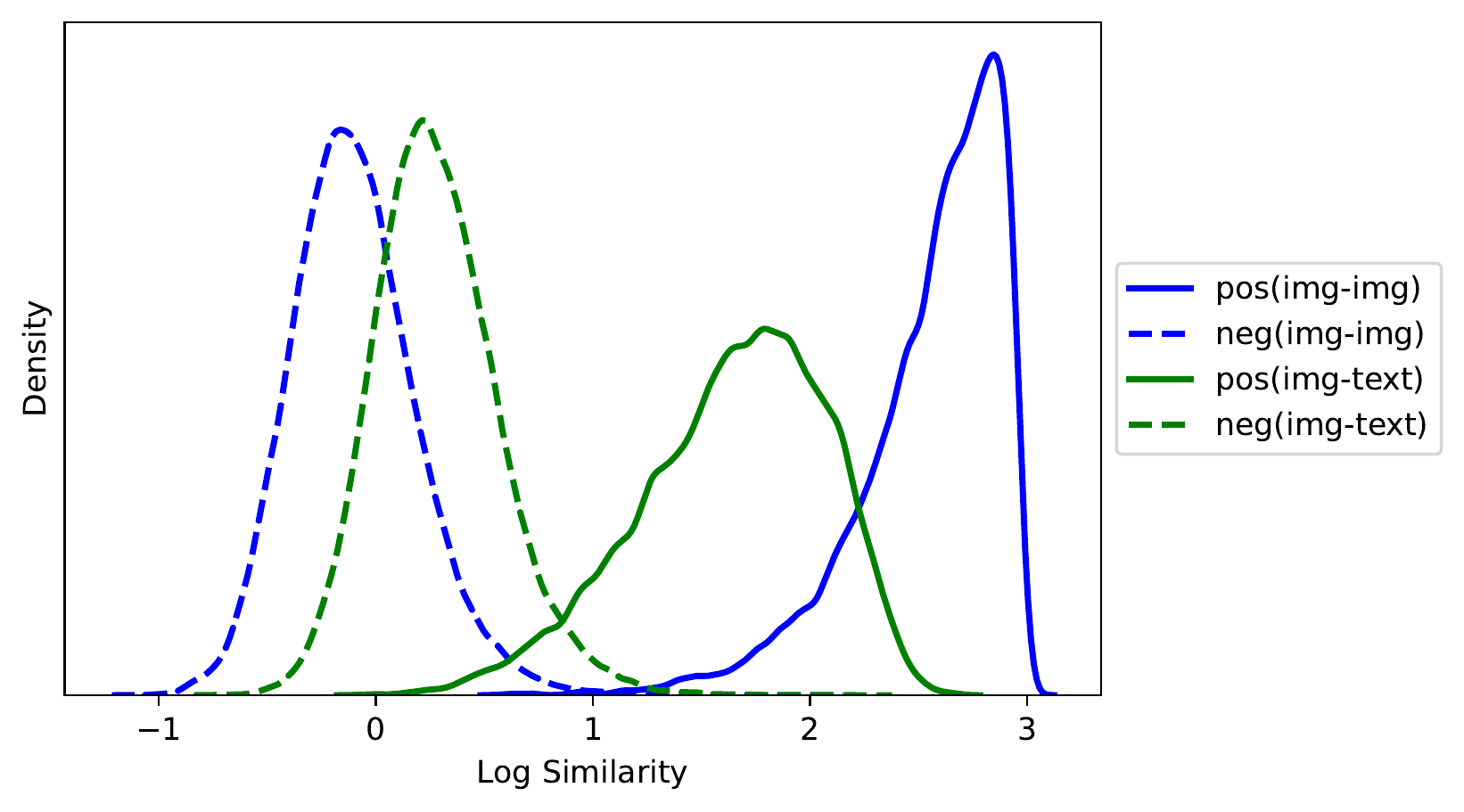}
        \caption{Domain-dependent $\tau, b$.}
        \label{fig:domain_dep}
    \end{subfigure}
    \caption{Density plot of similarity scores with respect to domain dependency of similarity measure.}
\end{figure}

\paragraph{Distribution of Similarities w.r.t Domain Dependency of Similarity Measure}
Figure~\ref{fig:shared} and Figure~\ref{fig:domain_dep} show distribution of similarities with respect to domain dependency of similarity measure. We can find that positives--negatives of image--image pairs separate better than positives--negatives of image--text pairs, which means the former pairs are easier to classify than the latter cases. With domain-dependent $\tau$ and $b$, those separations are more pronounced depending on the domain difficulties, resulting in better performance.

\section{Implementation Details}
\label{appendix:impdet}
\paragraph{Experimental Settings}
For the main experiments in Section \ref{main}, 
we used settings in Table~\ref{exp:config} for UniCLIP training. 
For the baselines, we used learning rate and weight decay of (5e-4, 2e-1) for CLIP~\cite{radford2021learning}, (3e-3, 1e-1) for SLIP~\cite{mu2021slip}, and (1e-3, 1e-1) for DeCLIP~\cite{cui2022democratizing}, while the remaining hyperparameters are the same as our method.
We used the implementation of CLIP and SLIP from \href{https://github.com/facebookresearch/SLIP}{https://github.com/facebookresearch/SLIP}, and DeCLIP from \href{https://github.com/Sense-GVT/DeCLIP}{https://github.com/Sense-GVT/DeCLIP}.
For a fair comparison, CLIP doubled the batch size to match memory usage.
All models are trained with the automatic mixed precision in PyTorch~\cite{paszke2019pytorch}.
Standard cropping and flipping augmentations~\cite{szegedy2015going} are used for linear probing, and RandAugment~\cite{cubuk2020randaugment} is used for fine-tuning.

\begin{table}[h]
    \caption{Training settings.}
    \label{exp:config}
    \centering
    \begin{adjustbox}{width=0.8\columnwidth,center}
    \begin{tabular}{ccccc}
        \toprule
        &  \multicolumn{2}{c}{Pre-train} & Linear probing & Fine-tuning \\
        Dataset $\rightarrow$ & YFCC15M & Open30M & 11 downstream & ImageNet \\
        Config $\downarrow$ \\
        \midrule
        Base learning rate &  1e-3 & 1e-3 & 1e-1 & 5e-4\\
        Weight decay & 0.2 & 0.1 & 0 & 0.05\\
        Epoch & 50 & 32 & 90 & 100\\
        Linear warmup epoch &  2 & 1 & 0 & 5\\
        Learning rate schedule & \multicolumn{2}{c}{Cosine decay} & - & - \\
        Optimizer & \multicolumn{2}{c}{AdamW} & SGD & AdamW\\
        Optimizer momentum & \multicolumn{2}{c}{0.9, 0.98}& 0.9 & 0.9, 0.999\\
        Total batch size & \multicolumn{2}{c}{4096} & 128 & 256 \\
        GPU & \multicolumn{2}{c}{16$\times$A100 40GB} & 1$\times$V100 16GB & 4$\times$V100 16GB \\
        \bottomrule
    \end{tabular}
    \end{adjustbox}
\end{table}

\paragraph{Augmentation Configurations}
Following our ablation studies in Tables \ref{exp:augtoken}
and \ref{abl:view}, one weakly augmented image view, two strongly augmented image views, and one text is used to train our networks.
Detailed image augmentation policies are described in Table~\ref{tab:aug_config}.
For text augmentations, EDA~\cite{wei2019eda} is applied only to CC3M since it has much more refined text data than other web-crawled noisy datasets like CC12M and YFCC15M. EDA is applied with a random replacement probability of 0.2 and a random deletion probability of 0.1.

\begin{table}[h]
    \caption{Image augmentation configurations in PyTorch style.}
    \centering
    \begin{adjustbox}{width=1.0\columnwidth,center}
    \begin{tabular}{ccccc}
        \toprule
         & Augmentation & Parameter & Value & Applying probability\\
        \midrule
        \multirow{3}{*}{Weak augmentation} & \texttt{RandomResizedCrop} & \texttt{size}, \texttt{scale}, \texttt{ratio} & 224, [0.5, 1], [3/4, 4/3] & 1\\
        & \texttt{ColorJitter} & \texttt{brightness}, \texttt{contrast}, \texttt{saturation}, \texttt{hue} & 0.4, 0.4, 0.4, 0.1 & 0.8\\
        & \texttt{GaussianBlur} & \texttt{kernel\_size}, \texttt{sigma} & 11, [0.1, 2] & 0.5\\
        \midrule
        \multirow{5}{*}{Strong augmentation} & \texttt{RandomResizedCrop} & \texttt{size}, \texttt{scale}, \texttt{ratio} & 224, [0.08, 1], [3/4, 4/3] & 1\\
        & \texttt{ColorJitter} & \texttt{brightness}, \texttt{contrast}, \texttt{saturation}, \texttt{hue} & 0.4, 0.4, 0.4, 0.1 & 0.8\\
        & \texttt{GaussianBlur} & \texttt{kernel\_size}, \texttt{sigma} & 11, [0.1, 2] & 0.5 \\
        & \texttt{RandomHorizontalFlip} & - & - & 0.5 \\
        & \texttt{RandomGrayscale} & - & - & 0.2 \\
        \midrule
        \multirow{5}{*}{Strong augmentation (DeCLIP)} & \texttt{RandomResizedCrop} & \texttt{size}, \texttt{scale}, \texttt{ratio} & 224, [0.2, 1], [3/4, 4/3] & 1\\
        & \texttt{ColorJitter} & \texttt{brightness}, \texttt{contrast}, \texttt{saturation}, \texttt{hue} & 0.4, 0.4, 0.4, 0.1 & 0.8\\
        & \texttt{GaussianBlur} & \texttt{kernel\_size}, \texttt{sigma} & 11, [0.1, 2] & 0.5 \\
        & \texttt{RandomHorizontalFlip} & - & - & 0.5 \\
        & \texttt{RandomGrayscale} & - & - & 0.2 \\
        \bottomrule
    \end{tabular}
    \end{adjustbox}
    \label{tab:aug_config}
\end{table}

\paragraph{Network Configurations}
Network configurations are summarized in Table~\ref{tab:network}.
The augmentation encoder is composed of 11-256-256-256 MLP with GELU activations.
A residual block in the projection head is identical to the feedforward module in Transformers and ViTs.
A linear layer follows 3 residual blocks in the projection head.

\begin{table}[h!]
    \centering
    \caption{Network configurations.}
    \label{tab:network}
    \begin{adjustbox}{width=0.8\columnwidth,center}
    \begin{tabular}{cccccccc}
        \toprule
        & &              & Input     & Output    & \multicolumn{3}{c}{Transformer}\\
        & & Architecture & dimension & dimension & layers & width & heads \\
        \midrule
        \multirow{3}{*}{Image} & Encoder $f_I$ & ViT-B/32 & 224$\times$224 & - & 12 & 768 & 12 \\
        & Augmentation encoder $f_A$ & 3-layer MLP & 11 & 256 & - & - & - \\
        & Projection head $g_I$ & 3 ResBlocks & 1024 & 512 & - & - & - \\
        \midrule
        \multirow{2}{*}{Text} & Encoder $f_T$ & Transformer & 77 & - & 12 & 512 & 8 \\
        & Projection head $g_T$ & Linear & 512 & 512 & - & - & - \\
        \bottomrule
    \end{tabular}
    \end{adjustbox}
\end{table}

\paragraph{Dataset Configurations}
Table~\ref{exp:dataset} describes all dataset configurations used in our experiments.
\begin{table}[h]
    \centering
    \caption{Dataset configurations. Flickr30k and COCO Captions have 5 captions per image.}
    \label{exp:dataset}
    % \begin{adjustbox}{width=1.0\columnwidth,center}
    \begin{tabular}{ccccc}
        \toprule
        & Dataset & \# Classes & \# Training & \# Validation (\# Test)\\
        \midrule
        \multirow{4}{*}{Pre-training}& CC3M & - & 2,891,358 & - \\
        & CC12M     & - & 10,663,994 & - \\
        & YFCC15M   & - & 15,171,110 & -\\
        & Open30M & - & 28,726,462 & -\\
        \midrule
        \multirow{11}{*}{\shortstack{Zero-shot classification \\ \& linear probing}} & Pets & 37 & 3,680 & 3,669\\
        & CIFAR-10 & 10 & 50,000 & 10,000\\
        & CIFAR-100 & 100 & 50,000 & 10,000\\
        & SUN397 & 397 & 19,850 & 19,850\\
        & Food-101 & 101 & 75,750 & 25,250 \\
        & Flowers & 102 & 2,040 & 6,149 \\
        & Cars & 196 & 8,144 & 8,041\\
        & Caltech-101 & 102 & 3,060 & 6,085\\
        & Aircraft & 100 & 6,667 & 3,333\\
        & DTD & 47 & 3,760 & 1,880\\
        & ImageNet & 1,000& 1,281,167 & 50,000\\
        \midrule
        \multirow{2}{*}{Image--text retrieval}& Flickr30k & - & 31,784 & 1,000 (1,000) \\
        & COCO Captions& - & 82,783 & 5,000 (5,000)\\
        \midrule
        \multirow{4}{*}{ImageNet variations} & ImageNet-R & 200 & - & 30,000\\
        & ImageNet-Sketch & 1,000 & - & 50,000 \\
        & ImageNetV2 & 1,000 & - & 30,000\\
        & ImageNet-A & 200 & - & 7,500 \\
        \bottomrule
    \end{tabular}
\end{table}

%% file: camera_ready.bbl
\begin{thebibliography}{47}
\providecommand{\natexlab}[1]{#1}
\providecommand{\url}[1]{\texttt{#1}}
\expandafter\ifx\csname urlstyle\endcsname\relax
  \providecommand{\doi}[1]{doi: #1}\else
  \providecommand{\doi}{doi: \begingroup \urlstyle{rm}\Url}\fi

\bibitem[Bossard et~al.(2014)Bossard, Guillaumin, and Gool]{bossard2014food}
L.~Bossard, M.~Guillaumin, and L.~V. Gool.
\newblock Food-101--mining discriminative components with random forests.
\newblock In \emph{European conference on computer vision}, pages 446--461.
  Springer, 2014.

\bibitem[Changpinyo et~al.(2021)Changpinyo, Sharma, Ding, and
  Soricut]{changpinyo2021conceptual}
S.~Changpinyo, P.~Sharma, N.~Ding, and R.~Soricut.
\newblock Conceptual 12m: Pushing web-scale image-text pre-training to
  recognize long-tail visual concepts.
\newblock In \emph{Proceedings of the IEEE/CVF Conference on Computer Vision
  and Pattern Recognition}, pages 3558--3568, 2021.

\bibitem[Chen et~al.(2020)Chen, Kornblith, Norouzi, and Hinton]{chen2020simple}
T.~Chen, S.~Kornblith, M.~Norouzi, and G.~Hinton.
\newblock A simple framework for contrastive learning of visual
  representations.
\newblock In \emph{International conference on machine learning}, pages
  1597--1607. PMLR, 2020.

\bibitem[Chen and He(2021)]{chen2021exploring}
X.~Chen and K.~He.
\newblock Exploring simple siamese representation learning.
\newblock In \emph{Proceedings of the IEEE/CVF Conference on Computer Vision
  and Pattern Recognition}, pages 15750--15758, 2021.

\bibitem[Chen et~al.(2015)Chen, Fang, Lin, Vedantam, Gupta, Doll{\'a}r, and
  Zitnick]{chen2015microsoft}
X.~Chen, H.~Fang, T.-Y. Lin, R.~Vedantam, S.~Gupta, P.~Doll{\'a}r, and C.~L.
  Zitnick.
\newblock Microsoft coco captions: Data collection and evaluation server.
\newblock \emph{arXiv preprint arXiv:1504.00325}, 2015.

\bibitem[Cimpoi et~al.(2014)Cimpoi, Maji, Kokkinos, Mohamed, and
  Vedaldi]{cimpoi2014describing}
M.~Cimpoi, S.~Maji, I.~Kokkinos, S.~Mohamed, and A.~Vedaldi.
\newblock Describing textures in the wild.
\newblock In \emph{Proceedings of the IEEE conference on computer vision and
  pattern recognition}, pages 3606--3613, 2014.

\bibitem[Cubuk et~al.(2020)Cubuk, Zoph, Shlens, and Le]{cubuk2020randaugment}
E.~D. Cubuk, B.~Zoph, J.~Shlens, and Q.~V. Le.
\newblock Randaugment: Practical automated data augmentation with a reduced
  search space.
\newblock In \emph{Proceedings of the IEEE/CVF Conference on Computer Vision
  and Pattern Recognition Workshops}, pages 702--703, 2020.

\bibitem[Cui et~al.(2022)Cui, Zhao, Liang, Li, and Shao]{cui2022democratizing}
Y.~Cui, L.~Zhao, F.~Liang, Y.~Li, and J.~Shao.
\newblock Democratizing contrastive language-image pre-training: A clip
  benchmark of data, model, and supervision.
\newblock \emph{arXiv preprint arXiv:2203.05796}, 2022.

\bibitem[Doersch and Zisserman(2017)]{doersch2017multi}
C.~Doersch and A.~Zisserman.
\newblock Multi-task self-supervised visual learning.
\newblock In \emph{Proceedings of the IEEE International Conference on Computer
  Vision}, pages 2051--2060, 2017.

\bibitem[Doersch et~al.(2015)Doersch, Gupta, and
  Efros]{doersch2015unsupervised}
C.~Doersch, A.~Gupta, and A.~A. Efros.
\newblock Unsupervised visual representation learning by context prediction.
\newblock In \emph{Proceedings of the IEEE international conference on computer
  vision}, pages 1422--1430, 2015.

\bibitem[Dosovitskiy et~al.(2021)Dosovitskiy, Beyer, Kolesnikov, Weissenborn,
  Zhai, Unterthiner, Dehghani, Minderer, Heigold, Gelly, Uszkoreit, and
  Houlsby]{dosovitskiy2021an}
A.~Dosovitskiy, L.~Beyer, A.~Kolesnikov, D.~Weissenborn, X.~Zhai,
  T.~Unterthiner, M.~Dehghani, M.~Minderer, G.~Heigold, S.~Gelly, J.~Uszkoreit,
  and N.~Houlsby.
\newblock An image is worth 16x16 words: Transformers for image recognition at
  scale.
\newblock In \emph{International Conference on Learning Representations}, 2021.

\bibitem[Fei-Fei et~al.(2004)Fei-Fei, Fergus, and Perona]{fei2004learning}
L.~Fei-Fei, R.~Fergus, and P.~Perona.
\newblock Learning generative visual models from few training examples: An
  incremental bayesian approach tested on 101 object categories.
\newblock In \emph{2004 conference on computer vision and pattern recognition
  workshop}, pages 178--178. IEEE, 2004.

\bibitem[Gidaris et~al.(2018)Gidaris, Singh, and
  Komodakis]{gidaris2018unsupervised}
S.~Gidaris, P.~Singh, and N.~Komodakis.
\newblock Unsupervised representation learning by predicting image rotations.
\newblock \emph{arXiv preprint arXiv:1803.07728}, 2018.

\bibitem[Hendrycks et~al.(2021{\natexlab{a}})Hendrycks, Basart, Mu, Kadavath,
  Wang, Dorundo, Desai, Zhu, Parajuli, Guo, et~al.]{hendrycks2021many}
D.~Hendrycks, S.~Basart, N.~Mu, S.~Kadavath, F.~Wang, E.~Dorundo, R.~Desai,
  T.~Zhu, S.~Parajuli, M.~Guo, et~al.
\newblock The many faces of robustness: A critical analysis of
  out-of-distribution generalization.
\newblock In \emph{Proceedings of the IEEE/CVF International Conference on
  Computer Vision}, pages 8340--8349, 2021{\natexlab{a}}.

\bibitem[Hendrycks et~al.(2021{\natexlab{b}})Hendrycks, Zhao, Basart,
  Steinhardt, and Song]{hendrycks2021natural}
D.~Hendrycks, K.~Zhao, S.~Basart, J.~Steinhardt, and D.~Song.
\newblock Natural adversarial examples.
\newblock In \emph{Proceedings of the IEEE/CVF Conference on Computer Vision
  and Pattern Recognition}, pages 15262--15271, 2021{\natexlab{b}}.

\bibitem[Jia et~al.(2021)Jia, Yang, Xia, Chen, Parekh, Pham, Le, Sung, Li, and
  Duerig]{jia2021scaling}
C.~Jia, Y.~Yang, Y.~Xia, Y.-T. Chen, Z.~Parekh, H.~Pham, Q.~Le, Y.-H. Sung,
  Z.~Li, and T.~Duerig.
\newblock Scaling up visual and vision-language representation learning with
  noisy text supervision.
\newblock In \emph{International Conference on Machine Learning}, pages
  4904--4916. PMLR, 2021.

\bibitem[Khosla et~al.(2020)Khosla, Teterwak, Wang, Sarna, Tian, Isola,
  Maschinot, Liu, and Krishnan]{khosla2020supervised}
P.~Khosla, P.~Teterwak, C.~Wang, A.~Sarna, Y.~Tian, P.~Isola, A.~Maschinot,
  C.~Liu, and D.~Krishnan.
\newblock Supervised contrastive learning.
\newblock \emph{Advances in Neural Information Processing Systems},
  33:\penalty0 18661--18673, 2020.

\bibitem[Kim and Ye(2021)]{kim2021diffusionclip}
G.~Kim and J.~C. Ye.
\newblock Diffusionclip: Text-guided image manipulation using diffusion models.
\newblock \emph{arXiv preprint arXiv:2110.02711}, 2021.

\bibitem[Krause et~al.(2013)Krause, Stark, Deng, and Fei-Fei]{krause20133d}
J.~Krause, M.~Stark, J.~Deng, and L.~Fei-Fei.
\newblock 3d object representations for fine-grained categorization.
\newblock In \emph{Proceedings of the IEEE international conference on computer
  vision workshops}, pages 554--561, 2013.

\bibitem[Krizhevsky et~al.(2009)Krizhevsky, Hinton,
  et~al.]{krizhevsky2009learning}
A.~Krizhevsky, G.~Hinton, et~al.
\newblock Learning multiple layers of features from tiny images.
\newblock 2009.

\bibitem[Li et~al.(2022)Li, Liang, Zhao, Cui, Ouyang, Shao, Yu, and
  Yan]{li2022supervision}
Y.~Li, F.~Liang, L.~Zhao, Y.~Cui, W.~Ouyang, J.~Shao, F.~Yu, and J.~Yan.
\newblock Supervision exists everywhere: A data efficient contrastive
  language-image pre-training paradigm.
\newblock In \emph{International Conference on Learning Representations}, 2022.

\bibitem[Maji et~al.(2013)Maji, Rahtu, Kannala, Blaschko, and
  Vedaldi]{maji2013fine}
S.~Maji, E.~Rahtu, J.~Kannala, M.~Blaschko, and A.~Vedaldi.
\newblock Fine-grained visual classification of aircraft.
\newblock \emph{arXiv preprint arXiv:1306.5151}, 2013.

\bibitem[Miech et~al.(2020)Miech, Alayrac, Smaira, Laptev, Sivic, and
  Zisserman]{miech2020end}
A.~Miech, J.-B. Alayrac, L.~Smaira, I.~Laptev, J.~Sivic, and A.~Zisserman.
\newblock End-to-end learning of visual representations from uncurated
  instructional videos.
\newblock In \emph{Proceedings of the IEEE/CVF Conference on Computer Vision
  and Pattern Recognition}, pages 9879--9889, 2020.

\bibitem[Mu et~al.(2021)Mu, Kirillov, Wagner, and Xie]{mu2021slip}
N.~Mu, A.~Kirillov, D.~Wagner, and S.~Xie.
\newblock Slip: Self-supervision meets language-image pre-training.
\newblock \emph{arXiv preprint arXiv:2112.12750}, 2021.

\bibitem[Nilsback and Zisserman(2008)]{nilsback2008automated}
M.-E. Nilsback and A.~Zisserman.
\newblock Automated flower classification over a large number of classes.
\newblock In \emph{2008 Sixth Indian Conference on Computer Vision, Graphics \&
  Image Processing}, pages 722--729. IEEE, 2008.

\bibitem[Noroozi and Favaro(2016)]{noroozi2016unsupervised}
M.~Noroozi and P.~Favaro.
\newblock Unsupervised learning of visual representations by solving jigsaw
  puzzles.
\newblock In \emph{European conference on computer vision}, pages 69--84.
  Springer, 2016.

\bibitem[Noroozi et~al.(2017)Noroozi, Pirsiavash, and
  Favaro]{noroozi2017representation}
M.~Noroozi, H.~Pirsiavash, and P.~Favaro.
\newblock Representation learning by learning to count.
\newblock In \emph{Proceedings of the IEEE International Conference on Computer
  Vision}, pages 5898--5906, 2017.

\bibitem[Parkhi et~al.(2012)Parkhi, Vedaldi, Zisserman, and
  Jawahar]{parkhi2012cats}
O.~M. Parkhi, A.~Vedaldi, A.~Zisserman, and C.~Jawahar.
\newblock Cats and dogs.
\newblock In \emph{2012 IEEE conference on computer vision and pattern
  recognition}, pages 3498--3505. IEEE, 2012.

\bibitem[Paszke et~al.(2019)Paszke, Gross, Massa, Lerer, Bradbury, Chanan,
  Killeen, Lin, Gimelshein, Antiga, et~al.]{paszke2019pytorch}
A.~Paszke, S.~Gross, F.~Massa, A.~Lerer, J.~Bradbury, G.~Chanan, T.~Killeen,
  Z.~Lin, N.~Gimelshein, L.~Antiga, et~al.
\newblock Pytorch: An imperative style, high-performance deep learning library.
\newblock \emph{Advances in neural information processing systems}, 32, 2019.

\bibitem[Pathak et~al.(2016)Pathak, Krahenbuhl, Donahue, Darrell, and
  Efros]{pathak2016context}
D.~Pathak, P.~Krahenbuhl, J.~Donahue, T.~Darrell, and A.~A. Efros.
\newblock Context encoders: Feature learning by inpainting.
\newblock In \emph{Proceedings of the IEEE conference on computer vision and
  pattern recognition}, pages 2536--2544, 2016.

\bibitem[Plummer et~al.(2015)Plummer, Wang, Cervantes, Caicedo, Hockenmaier,
  and Lazebnik]{plummer2015flickr30k}
B.~A. Plummer, L.~Wang, C.~M. Cervantes, J.~C. Caicedo, J.~Hockenmaier, and
  S.~Lazebnik.
\newblock Flickr30k entities: Collecting region-to-phrase correspondences for
  richer image-to-sentence models.
\newblock In \emph{Proceedings of the IEEE international conference on computer
  vision}, pages 2641--2649, 2015.

\bibitem[Radford et~al.(2021)Radford, Kim, Hallacy, Ramesh, Goh, Agarwal,
  Sastry, Askell, Mishkin, Clark, et~al.]{radford2021learning}
A.~Radford, J.~W. Kim, C.~Hallacy, A.~Ramesh, G.~Goh, S.~Agarwal, G.~Sastry,
  A.~Askell, P.~Mishkin, J.~Clark, et~al.
\newblock Learning transferable visual models from natural language
  supervision.
\newblock In \emph{International Conference on Machine Learning}, pages
  8748--8763. PMLR, 2021.

\bibitem[Ramesh et~al.(2021)Ramesh, Pavlov, Goh, Gray, Voss, Radford, Chen, and
  Sutskever]{ramesh2021zero}
A.~Ramesh, M.~Pavlov, G.~Goh, S.~Gray, C.~Voss, A.~Radford, M.~Chen, and
  I.~Sutskever.
\newblock Zero-shot text-to-image generation.
\newblock In \emph{International Conference on Machine Learning}, pages
  8821--8831. PMLR, 2021.

\bibitem[Recht et~al.(2019)Recht, Roelofs, Schmidt, and
  Shankar]{recht2019imagenet}
B.~Recht, R.~Roelofs, L.~Schmidt, and V.~Shankar.
\newblock Do imagenet classifiers generalize to imagenet?
\newblock In \emph{International Conference on Machine Learning}, pages
  5389--5400. PMLR, 2019.

\bibitem[Russakovsky et~al.(2015)Russakovsky, Deng, Su, Krause, Satheesh, Ma,
  Huang, Karpathy, Khosla, Bernstein, et~al.]{russakovsky2015imagenet}
O.~Russakovsky, J.~Deng, H.~Su, J.~Krause, S.~Satheesh, S.~Ma, Z.~Huang,
  A.~Karpathy, A.~Khosla, M.~Bernstein, et~al.
\newblock Imagenet large scale visual recognition challenge.
\newblock \emph{International journal of computer vision}, 115\penalty0
  (3):\penalty0 211--252, 2015.

\bibitem[Schroff et~al.(2015)Schroff, Kalenichenko, and
  Philbin]{schroff2015facenet}
F.~Schroff, D.~Kalenichenko, and J.~Philbin.
\newblock Facenet: A unified embedding for face recognition and clustering.
\newblock In \emph{Proceedings of the IEEE conference on computer vision and
  pattern recognition}, pages 815--823, 2015.

\bibitem[Sharma et~al.(2018)Sharma, Ding, Goodman, and
  Soricut]{sharma2018conceptual}
P.~Sharma, N.~Ding, S.~Goodman, and R.~Soricut.
\newblock Conceptual captions: A cleaned, hypernymed, image alt-text dataset
  for automatic image captioning.
\newblock In \emph{Proceedings of the 56th Annual Meeting of the Association
  for Computational Linguistics (Volume 1: Long Papers)}, pages 2556--2565,
  2018.

\bibitem[Sohn(2016)]{sohn2016improved}
K.~Sohn.
\newblock Improved deep metric learning with multi-class n-pair loss objective.
\newblock \emph{Advances in neural information processing systems}, 29, 2016.

\bibitem[Szegedy et~al.(2015)Szegedy, Liu, Jia, Sermanet, Reed, Anguelov,
  Erhan, Vanhoucke, and Rabinovich]{szegedy2015going}
C.~Szegedy, W.~Liu, Y.~Jia, P.~Sermanet, S.~Reed, D.~Anguelov, D.~Erhan,
  V.~Vanhoucke, and A.~Rabinovich.
\newblock Going deeper with convolutions.
\newblock In \emph{Proceedings of the IEEE conference on computer vision and
  pattern recognition}, pages 1--9, 2015.

\bibitem[Thomee et~al.(2016)Thomee, Shamma, Friedland, Elizalde, Ni, Poland,
  Borth, and Li]{thomee2016yfcc100m}
B.~Thomee, D.~A. Shamma, G.~Friedland, B.~Elizalde, K.~Ni, D.~Poland, D.~Borth,
  and L.-J. Li.
\newblock Yfcc100m: The new data in multimedia research.
\newblock \emph{Communications of the ACM}, 59\penalty0 (2):\penalty0 64--73,
  2016.

\bibitem[Van~den Oord et~al.(2018)Van~den Oord, Li, and
  Vinyals]{van2018representation}
A.~Van~den Oord, Y.~Li, and O.~Vinyals.
\newblock Representation learning with contrastive predictive coding.
\newblock \emph{arXiv e-prints}, pages arXiv--1807, 2018.

\bibitem[Vaswani et~al.(2017)Vaswani, Shazeer, Parmar, Uszkoreit, Jones, Gomez,
  Kaiser, and Polosukhin]{vaswani2017attention}
A.~Vaswani, N.~Shazeer, N.~Parmar, J.~Uszkoreit, L.~Jones, A.~N. Gomez,
  {\L}.~Kaiser, and I.~Polosukhin.
\newblock Attention is all you need.
\newblock \emph{Advances in neural information processing systems}, 30, 2017.

\bibitem[Wang et~al.(2019)Wang, Ge, Lipton, and Xing]{wang2019learning}
H.~Wang, S.~Ge, Z.~Lipton, and E.~P. Xing.
\newblock Learning robust global representations by penalizing local predictive
  power.
\newblock \emph{Advances in Neural Information Processing Systems}, 32, 2019.

\bibitem[Wei and Zou(2019)]{wei2019eda}
J.~Wei and K.~Zou.
\newblock Eda: Easy data augmentation techniques for boosting performance on
  text classification tasks.
\newblock In \emph{Proceedings of the 2019 Conference on Empirical Methods in
  Natural Language Processing and the 9th International Joint Conference on
  Natural Language Processing (EMNLP-IJCNLP)}, pages 6382--6388, 2019.

\bibitem[Xiao et~al.(2016)Xiao, Ehinger, Hays, Torralba, and
  Oliva]{xiao2016sun}
J.~Xiao, K.~A. Ehinger, J.~Hays, A.~Torralba, and A.~Oliva.
\newblock Sun database: Exploring a large collection of scene categories.
\newblock \emph{International Journal of Computer Vision}, 119\penalty0
  (1):\penalty0 3--22, 2016.

\bibitem[Zhang et~al.(2016)Zhang, Isola, and Efros]{zhang2016colorful}
R.~Zhang, P.~Isola, and A.~A. Efros.
\newblock Colorful image colorization.
\newblock In \emph{European conference on computer vision}, pages 649--666.
  Springer, 2016.

\bibitem[Zhang et~al.(2017)Zhang, Isola, and Efros]{zhang2017split}
R.~Zhang, P.~Isola, and A.~A. Efros.
\newblock Split-brain autoencoders: Unsupervised learning by cross-channel
  prediction.
\newblock In \emph{Proceedings of the IEEE Conference on Computer Vision and
  Pattern Recognition}, pages 1058--1067, 2017.

\end{thebibliography}
